\documentclass[10pt]{book}

\usepackage[table]{xcolor}
\usepackage{mathrsfs} % for mathscr
\usepackage{amssymb,amsmath}
\usepackage{graphicx}
\usepackage{hyperref} 
\usepackage{xspace}
\usepackage{booktabs}
\usepackage{caption} 
\usepackage{longtable}

\usepackage{pdfpages}
\usepackage{caption}

\hypersetup{colorlinks,linkcolor={red!50!black},citecolor={blue!50!black},urlcolor={blue!80!black}}
\usepackage[top=1.2in, bottom=1.2in, left=1.4in, right=1.4in]{geometry}

\usepackage{tikz}
\usetikzlibrary{shapes,calc,positioning,automata,arrows,trees}

\newcommand\bcpen{\includegraphics[width=15pt]{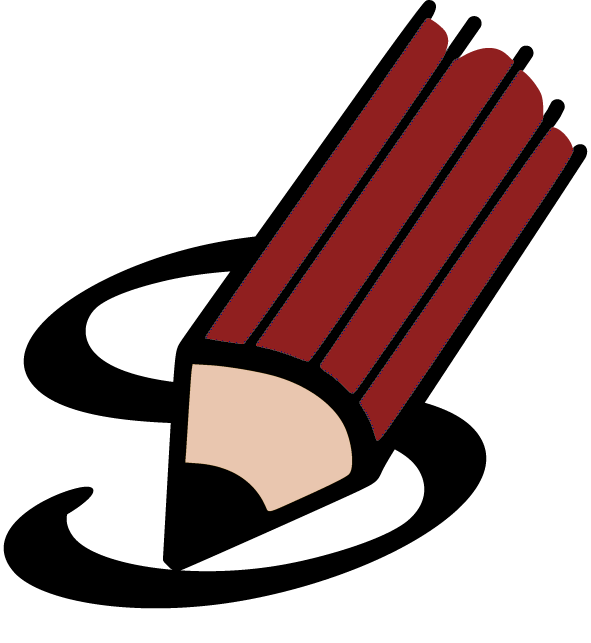}}

\usepackage{listingsutf8}
\usepackage{textcomp}
\usepackage{verbatim}
\usepackage{multirow}

\definecolor{v3lgray}{gray}{0.98}
\definecolor{v2lgray}{gray}{0.85}
\definecolor{vlgray}{gray}{0.92}
\definecolor{dgray}{rgb}{0.4,0.4,0.4}
\definecolor{dblue}{RGB}{0,0,99}
\definecolor{dred}{RGB}{175,6,54}
\definecolor{lred}{RGB}{155,6,44}
\definecolor{dgreen}{RGB}{47,135,7}
\definecolor{dviolet}{RGB}{102,0,153}
\definecolor{mblue}{RGB}{0,0,180}
\definecolor{dorange}{RGB}{204, 82, 0}

\def\ace{ACE\xspace}

\def\x3{{\rm XCSP$^3$}\xspace}
\def\jv3{{\rm JvCSP$^3$}\xspace}
\def\p3{{\rm PyCSP$^3$}\xspace}

\newcommand{\gb}[1]{{\tt #1}} % global constraint names
\lstdefinelanguage{json}{
    basewidth  = {.6em,0.6em},
    basicstyle=\normalfont\ttfamily,
    breaklines=true,
    morestring=[b]',
    morestring=[b]", 
    sensitive=false,
    stringstyle=\color[rgb]{0.227,0.226,0.441}\ttfamily, 
    escapechar=!,
    showstringspaces=false,
    xleftmargin=20pt, %,xrightmargin=-20pt,
    breaklines=true,basicstyle=\ttfamily\small,inputencoding=utf8/latin9,texcl
}
\lstnewenvironment{json}{\lstset{language=json}}{}

\lstdefinelanguage{mcsp}{
  keywords={forall,array,block,class,implements,model,public,slide},
  basewidth  = {.6em,0.6em},
  keywordstyle=\color{dred}\bfseries,
  ndkeywords={intension,lessThan,lessEqual,greaterEqual,greaterThan,equal,different,implication,equivalence,conjunction,disjunction,extension,regular,mdd,allDifferent,allDifferentMatrix,allEqual,ordered,increasing,decreasing,strictlyIncreasing,strictlyDecreasing,lex,lexMatrix,sum,count,atMost,atLeast,exactly,atMost1,atleast1,exactly1,element,channel,maximum,minimum,cardinality,nValues,noOverlap,cumulative,instantiation,clause,circuit,minimize,maximize},
  ndkeywordstyle=\color{mblue}\bfseries,
  identifierstyle=\color{black},
  sensitive=false,
  comment=[l]{//},
  morecomment=[s]{<!--}{-->},
  commentstyle=\color{dgreen}\ttfamily,
  stringstyle=\color{dgreen}\rmfamily, %normalfont,
  morestring=[b]',
  morestring=[b]",
  escapechar=~,
  showstringspaces=false,
  classoffset=2, morekeywords={private},keywordstyle=\color{gray},
  classoffset=3, morekeywords={dom,size,when},keywordstyle=\color{dorange},
  xleftmargin=-22pt,xrightmargin=-22pt,
  breaklines=true,basicstyle=\ttfamily\footnotesize,backgroundcolor=\color{v3lgray},inputencoding=utf8/latin9,texcl
}
\lstnewenvironment{mcsp}{\lstset{language=mcsp}}{}

\definecolor{colorex}{RGB}{255,248,220} %{HTML}{EDEDED}
\definecolor{mgray}{rgb}{0.55,0.55,0.55}
\definecolor{officegreen}{rgb}{0.0, 0.5, 0.0}

\newcounter{cntPy}

\lstnewenvironment{python}{\lstset{
    inputencoding=utf8/latin1,
    language=python,
    ndkeywords={size,dom,matrix,strict,occurrences},
    ndkeywordstyle=\color{mblue}, %NavyBlue}, %brickred}, %NavyBlue},
    keywords=[2]{data,satisfy,minimize,maximize,solve}, % ,If,Then,Else},
    keywordstyle=[2]\color{dred}, %NavyBlue}, %brickred}, %NavyBlue},
    deletekeywords={def},
    keywords=[3]{def,If,Then,Match}, %def},
    keywordstyle=[3]\color{dorange},
    commentstyle=\color{officegreen}, %OliveGreen},
    showstringspaces=false,
    captionpos=t,
    basicstyle=\ttfamily\footnotesize,
    framesep=5pt,
    xleftmargin=5pt,xrightmargin=0pt,
    backgroundcolor=\color{colorex}, %rnsilk}, %OliveGreen},deepchampagne},
    escapechar=@% char to escape out of listings and back to LaTeX
  }
}{}

\usepackage[skins,breakable,xparse]{tcolorbox}
\newcommand{\core}[1]{ 
  \medskip \begin{tcolorbox}[
    enhanced,breakable,
    boxsep=0pt,top=4pt,bottom=0pt,left=2mm,right=1mm,
    toprule=0.1mm,leftrule=0.1mm,rightrule=0.25mm,bottomrule=0.25mm,shadow={0.2mm}{-0.2mm}{0mm}{dgray},
    overlay unbroken and first={\node (logo) at ([xshift=4mm,yshift=-5mm]frame.north west) {#1}; },
    colframe=dgray,titlerule=-0.2mm,toptitle=3mm,coltitle=dred, fonttitle=\bfseries,
    lines before break=6, pad at break*=10pt
    }

\newenvironment{boxpy}
 {\stepcounter{cntPy} \core{\bcpen} , colback=colorex, title style={color=colorex}, title=~ ~ \,PyCSP$^3$ Model \thecntPy]}
 {\end{tcolorbox}}

\newenvironment{command}
  {\quote\small\verbatim}
  {\endverbatim\endquote}

  \usepackage{authblk}
 
\title{\textcolor{dred}{Proceedings of the \x3 Competition 2023}}
\author{Gilles Audemard \and Christophe Lecoutre \and Emmanuel Lonca}
\affil{\vspace{1cm} CRIL \\University of Artois \& CNRS \\ France}

\date{August 30, 2023 \\ {\small (Revised$^*$ on December 10, 2023)}}

\begin{document}
\maketitle

~ \\
~ \\

\bigskip

This document represents the proceedings of the \x3 Competition 2023.
%This is a preliminary version containing the description of problems (models) selected for the competition.
%Soon, it will also contain the description of the solvers and the results.
%\bigskip
The website containing all {\bf detailed results} of this international competition is available at:
\begin{quote}
  \href{https://www.cril.univ-artois.fr/XCSP23/}{https://www.cril.univ-artois.fr/XCSP23}
\end{quote}

  \bigskip
  \bigskip
\noindent The organization of this 2023 competition involved the following tasks:
\begin{itemize}
\item adjusting general details (dates, tracks, $\dots$) by G. Audemard, C. Lecoutre and E. Lonca
\item selecting instances (problems, models and data) by C. Lecoutre
\item receiving, testing and executing solvers on CRIL cluster by E. Lonca
\item validating solvers and rankings by C. Lecoutre and  E. Lonca
\item developping the 2023 website dedicated to results by G. Audemard
\end{itemize}

\bigskip\bigskip 
{\bf Important}: for reproducing the experiments and results, it is important to use the set of \x3 instances used in the competition.
These instances can be found in this \href{https://www.cril.univ-artois.fr/~lecoutre/compets/instancesXCSP23.zip}{archive}.
Some (usually minor) differences may exist when compiling the models presented in this document and those that can be found in this  \href{https://www.cril.univ-artois.fr/~lecoutre/compets/modelsXCSP23.zip}{archive}.

\bigskip
    {\bf Revision} ($^*$) of December 2023: some models in this document have been simplified while using new possibilities offered by Version 2.2 of \p3.
As mentioned above, note that in order to reproduce results and/or to make fair new comparisons with respect to solvers engaged in the 2023 competition, you have to use the very \href{https://www.cril.univ-artois.fr/~lecoutre/compets/instancesXCSP23.zip}{same set} of \x3 instances, as in the 2023 competition.

\tableofcontents

\chapter{About the Selection of Problems in 2023}

Remember that the complete description, {\bf Version 3.1}, of the format (\x3) used to represent combinatorial constrained problems can be found in \cite{BLAPxcsp3}.
As usual for \x3 competitions, we have limited \x3 to its kernel, called \x3-core \cite{BLAP_xcsp3core}.
This means that the scope of the \x3 competition is restricted to:
\begin{itemize}
\item integer variables,
\item CSP (Constraint Satisfaction Problem) and COP (Constraint Optimization Problem), 
\item a set of 24 popular (global) constraints for main tracks:
  \begin{itemize}
  \item generic constraints: \gb{intension} and \gb{extension} (also called \gb{table})
  \item language-based constraints: \gb{regular} and \gb{mdd}
  \item comparison constraints: \gb{allDifferent}, \gb{allEqual}, \gb{ordered}, \gb{lex} and \gb{precedence}
  \item counting/summing constraints: \gb{sum}, \gb{count}, \gb{nValues} and \gb{cardinality}
  \item connection constraints: \gb{maximum}, \gb{minimum}, \gb{element} and \gb{channel}
  \item packing/scheduling constraints: \gb{noOverlap}, \gb{cumulative}, \gb{binPacking} and \gb{knapsack}
  \item \gb{circuit}, \gb{instantiation} and \gb{slide}
  \end{itemize}
  and a small set of constraints for mini tracks.
\end{itemize}

Note that \x3-core has been extended in Version 3.1 so as to accept \gb{precedence}, \gb{binPacking} and \gb{knapsack}.

\medskip
For the 2023 competition, 36 problems have been selected.
They are succinctly presented in Table \ref{fig:problems}.
For each problem, the type of optimization is indicated (if any), as well as the involved constraints.
At this point, do note that making a good selection of problems/instances is a difficult task.
In our opinion, important criteria for a good selection are:
\begin{itemize}
\item the novelty of problems, avoiding constraint solvers to overfit already published problems;
\item the diversity of constraints, trying to represent all of the most popular constraints (those from \x3-core) while paying attention to not over-representing some of them;
\item the scaling up of problems.
\end{itemize}

\begin{table}
  \begin{small}
  \begin{tabular}{p{4cm}p{3cm}p{7.5cm}}
    \toprule
CSP Problems & & Global Constraints\\
    \midrule
\rowcolor{vlgray}{}  AnotherMagicSquare & & \gb{allDifferent}, \gb{sum}  \\
    AntimagicSquare &   & \gb{allDifferent}, \gb{maximum}, \gb{minimum}, \gb{sum} \\
\rowcolor{vlgray}{}  BinaryPuzzle   &  &\gb{regular},  \gb{sum}, \gb{table} ($*$) \\
    CalvinPuzzle & & \gb{allDifferent}, \gb{count}, \gb{table} ($*$) \\
\rowcolor{vlgray}{}   Coloring & &  \\
    CoveringArray & & \gb{allDifferent}, \gb{channel}, \gb{table} \\
\rowcolor{vlgray}{}  Dominoes & & \gb{allDifferent}, \gb{table}\\
Fischer & &  \\
\rowcolor{vlgray}{} MagicSquare & & \gb{allDifferent}, \gb{mdd}, \gb{sum} \\
NonogramTernary & & \gb{table} \\
\rowcolor{vlgray}{} NonTransitiveDice & &  \gb{maximum}, \gb{sum} \\
PegSolitaire &  &  \\
\rowcolor{vlgray}{} Primes & & \gb{sum} \\
PythagoreanTriples &  & \gb{nValues} \\
\rowcolor{vlgray}{} Slant & & \gb{count}, \gb{sum}  \\
Soccer &  & \gb{allDifferent}, \gb{sum}, \gb{table} \\
\rowcolor{vlgray}{} SquarePackingSuite & & \gb{cumulative}, \gb{noOverlap} \\
WordDesignDNA &  &  \gb{mdd}, \gb{lex}, \gb{sum},  \gb{table} \\
%\rowcolor{vlgray}{} Solitaire Batleship & & \gb{cardinality}, \gb{count}, \gb{extension}, \gb{regular}, \gb{sum} \\
%Sports Scheduling & & \gb{allDifferent}, \gb{cardinality}, \gb{count}, \gb{extension} \\
%\rowcolor{vlgray}{} Superpermutation & & \gb{allDifferent}, \gb{element} \\
%\midrule
& & \\
\midrule
COP Problems & Optimization & Global Constraints\\
\midrule
\rowcolor{vlgray}{} AircraftAssemblyLine &  $\min$ SUM & \gb{cumulative}, \gb{noOverlap} \\
 BeerJugs &  $\max$ VAR & \gb{table} \\
\rowcolor{vlgray}{} Benzenoide &  $\min$ SUM & \gb{count}, \gb{lex}, \gb{precedence}, \gb{sum}, \gb{table} ($*$) \\
CarpetCutting & $\min$ VAR & \gb{cumulative}, \gb{element}, \gb{noOverlap}, \gb{table} \\
\rowcolor{vlgray}{} GBAC &  $\min$ SUM & \gb{binPacking}, \gb{cardinality}, \gb{maximum}, \gb{table} \\
GeneralizedMKP & $\max$ VAR & \gb{knapsack}, \gb{sum} \\
\rowcolor{vlgray}{} HCPizza & $\max$ VAR & \gb{sum}, \gb{table} \\
HSP & $\min$ VAR & \gb{allDifferent}, \gb{maximum}, \gb{noOverlap}, \gb{table}  \\
\rowcolor{vlgray}{} KidneyExchange & $\max$ SUM & \gb{allDifferent}, \gb{binPacking}, \gb{element}, \gb{precedence} \\
KMedian &  $\min$ SUM & \gb{allDifferent}, \gb{element}, \gb{minimum}, \gb{sum} \\
\rowcolor{vlgray}{} LargeScaleScheduling & $\min$ MAXIMUM & \gb{cumulative}, \gb{maximum} \\
ProgressiveParty & $\min$ SUM & \gb{allDifferent}, \gb{channel}, \gb{element}, \gb{sum} \\
\rowcolor{vlgray}{} PSP & $\min$ SUM &  \gb{count}, \gb{element}, \gb{sum}  \\
RIP & $\min$ SUM &  \gb{cumulative}\\
\rowcolor{vlgray}{} RuleMining &  $\max$ VAR & \gb{allDifferent}, \gb{count}, \gb{table} ($*$) \\
Sonet & $\min$ SUM &\gb{lex},  \gb{sum}, \gb{table} ($*$) \\
\rowcolor{vlgray}{} SRFLP & $\min$ SUM & \gb{allDifferent}, \gb{sum}, \gb{table} \\
TSPTW & $\min$ SUM & \gb{allDifferent},  \gb{circuit}, \gb{element} \\
%\rowcolor{vlgray}{} War or Peace & $\min$ SUM & \gb{sum} \\
\bottomrule
  \end{tabular}
  \end{small}
  \caption{Selected Problems for the main tracks of the 2023 Competition. VAR/EXPR means that a variable/expression must be optimized. When \gb{extension} is followed by ($*$), it means that starred tables are involved.}\label{fig:problems}
\end{table}

\paragraph{Novelty.} Almost all problems are new in 2023, with models directly written in \p3. Three problems have been submitted, in response to the call. 
%ore than one third of the problems are new (15 out of 41, i.e. 36.5\%). They are Auction, BACP, Crosswords Design, Eternity, FAPP, Mistery Shopper, Nurse Rostering, Peaceable Armies, Pizza Voucher, RLFAP, Steel Mill Slab, Sum Coloring, TAL, Template Design, and Traveling Tournament.
%Some of these new problems are quite challenging; in particular, Crosswords design (an optimization problem with a total freedom on the position of black cells), FAPP (the original optimization instances from the ROADEF challenge), Nurse Rostering (an optimization problem involving many types of constraints), RLFAP (the original optimization instances from the ``Centre d'Electronique de l'Armement'') and TAL (an optimization problem of natural language processing).
%It is important to note that the optimization FAPP and RLFAP instances, mentioned here, are far more difficult that the simplified satisfaction versions published in XCSP 2.1 some years ago. 

%\paragraph{Diversity.} Of course, not all types of constraints are equally involved in the selected benchmark.
%In the next edition, we shall attempt to foster the representation of CP-representative global constraints such as \gb{cumulative} and \gb{noOverlap}, and possibly, to introduce a couple of new ones (e.g., \gb{binPacking}).

\paragraph{Scaling up.} It is always interesting to see how constraint solvers behave when the instances of a problem become harder and harder.
This is what we call the scaling behavior of solvers.
For most of the problems in the 2023 competition, we have selected series of instances with regular increasing difficulty.
It is important to note that assessing the difficulty of instances was mainly determined with \ace \cite{ace}, which is the reason why \ace is declared to be off-competition (due to this strong bias).

\paragraph{Selection.} This year, the selection of problems and instances has been performed by Christophe Lecoutre.
As a consequence, the solver \ace was labeled off-competition.

%\section{About the models}

\bigskip

\chapter{Problems and Models}

In the next sections, you will find all models used for generating the \x3 instances of the 2023 competition (for main CSP and COP tracks).
Almost all models are written in \p3 \cite{LS_pycsp3}, Version 2.1, officially released in November 2022; see \href{https://pycsp.org}{https://pycsp.org}.

\section{CSP}

\subsection{Another Magic Square}

\paragraph{Description.}
This puzzle is defined at the ``Fun with num3ers'' website; see \href{http://benvitale-funwithnum3ers.blogspot.com/2010/12/another-kind-of-magic-square.html}{benvitale-funwithnum3ers.blogspot.com}.
On a square grid of size $n \times n$, all numbers ranging from 1 to $n^2$ must be put so that the numbers surrounding each number add to a multiple of that number.

\paragraph{Data.}

Only one integer is required to specify a specific instance: the order $n$ of the grid. The values of $n$ used for generating the 2023 competition instances are:
\begin{quote}
2, 3, 4, 5, 6, 7, 8, 9, 10, 12 
\end{quote}

\paragraph{Model.}
The \p3 model, in a file `AnotherMagicSquare.py', used for the competition is:

\begin{boxpy}\begin{python}
@\imp@

n = data

# x[i][j] is the value at row i and column j
x = VarArray(size=[n, n], dom=range(1, n * n + 1))

satisfy(
   AllDifferent(x),

   # ensuring that the numbers surrounding a number v add to a multiple of v
   [Sum(x.around(i, j)) % x[i][j] == 0 for i in range(n) for j in range(n)]
)
\end{python}\end{boxpy}

The model involves a two-dimensional array of variable $x$, a constraint \gb{AllDifferent} and a group of constraints \gb{Sum}.
A series of 10 instances has been selected for the competition.
For generating an \x3 instance (file), you can execute for example:
\begin{command}
python AnotherMagicSquare.py -data=10
\end{command}

\subsection{Antimagic Square}

\paragraph{Description.}
An antimagic square of order $n$ is an arrangement of the numbers 1 to $n^2$ in a square, such that the sums of the $n$ rows, the $n$ columns and the two diagonals form a sequence of $2n + 2$ consecutive integers; see \href{https://en.wikipedia.org/wiki/Antimagic_square}{wikipedia}.

\paragraph{Data.}
Only one integer is required to specify a specific instance: the order $n$ of the square. The values of $n$ used for generating the 2023 competition instances are: 
\begin{quote}
3, 4, 5, 6, 7, 8, 9, 10, 11, 12 
\end{quote}

\paragraph{Model.}
The \p3 model, in a file `AntimagicQuare.py', used for the competition is:

\begin{boxpy}\begin{python}
@\imp@

n = data

lb, ub = (n * (n + 1)) // 2, ((n * n) * (n * n + 1)) // 2

# x[i][j] is the value put in the cell of the matrix at coordinates (i,j)
x = VarArray(size=[n, n], dom=range(1, n * n + 1))

# y[k] is the sum of values in the kth line (row, column or diagonal)
y = VarArray(size=2 * n + 2, dom=range(lb, ub + 1))

satisfy(
   # all values must be different
   AllDifferent(x),

   # computing sums
   [
      [y[i] == Sum(x[i]) for i in range(n)],
      [y[n + j] == Sum(x[:, j]) for j in range(n)],
      y[2 * n] == Sum(diagonal_down(x)),
      y[2 * n + 1] == Sum(diagonal_up(x))
   ],

   # all sums must be consecutive
   [
      AllDifferent(y),
      Maximum(y) - Minimum(y) == 2 * n + 1
   ],

   # tag(symmetry-breaking)
   # ensuring Frenicle standard form
   [
      x[0][0] < x[0][-1],
      x[0][0] < x[-1][0],
      x[0][0] < x[-1][-1],
      x[0][1] < x[1][0]
   ]
)
\end{python}\end{boxpy}

The model involves two arrays of variables $x$ and $y$; the second one allows us to compute/record the sums of the different lines.
Consecutivness is ensured by a constraint involving \gb{Maximum} and \gb{Minimum}.
A series of 10 instances has been selected for the competition.
For generating an \x3 instance (file), you can execute for example:
\begin{command}
python AntimagicSquare.py -data=10
\end{command}

\subsection{Binary Puzzle}

\paragraph{Description.}
A binary puzzle (also known as a binary Sudoku) is a puzzle played on a $n \times n$ grid; initially some of the cells may contain 0 or 1 (but this is not the case for the 2023 competition).
One has to fill the remaining empty cells with either 0 or 1 according to the following rules:
\begin{itemize}
\item no more than two similar numbers next to or below each other are allowed,
\item each row and each column should contain an equal number of zeros and ones,
\item each row is unique and each column is unique.
\end{itemize}
See \cite{D_binary}.

\paragraph{Data.}

Only one integer is required to specify a specific instance: the order $n$ of the grid. The values of $n$ used for generating the 2023 competition instances are: 
\begin{quote}
20, 40, 60, 80, 100, 120
\end{quote}

\paragraph{Model.}

The \p3 model, in a file `BinaryPuzzle.py', used for the competition is:

\begin{boxpy}\begin{python}
@\imp@

n = data
assert n % 2 == 0
m = n // 2

# x[i][j] is the value in the cell of the grid at coordinates (i,j)
x = VarArray(size=[n, n], dom={0, 1})

if not variant():
   satisfy(
      # ensuring the same number of 0s and 1s in rows
      [Sum(x[i]) == m for i in range(n)],

      # ensuring the same number of 0s and 1s in columns
      [Sum(x[:, j]) == m for j in range(n)],
      
      # forbidding sequences of 3 consecutive 0s or 1s in rows
      [Sum(x[i, j:j + 3]) in {1,2} for i in range(n) for j in range(n - 2)],
      
      # forbidding sequences of 3 consecutive 0s or 1s in columns
      [Sum(x[i:i + 3, j]) in {1,2} for j in range(n) for i in range(n - 2)]
   )

elif variant("regular"):
   pairs = [(j, k) for j in range(3) for k in range(3) if (j==0 and k>0) or (j>0 and k==0)]

   q = Automaton.q
   t = [(q(0,0,0), 0, q(0,1,0)), (q(0,0,0), 1, (q(1,0,1)))] 
       + [(q(i,j,k), 0, q(i,j+1,0)) for i in range(m + 1) for j, k in pairs if j < 2] 
       + [(q(i,j,k), 1, q(i+1,0,k+1)) for i in range(m) for j, k in pairs if k < 2]
   A = Automaton(start=q(0,0,0), final=[q(m,j,k) for j, k in pairs], transitions=t)

   satisfy(
      # ensuring valid rows 
      [x[i] in A for i in range(n)],

      # ensuring valid columns 
      [x[:, j] in A for j in range(n)]
   )

satisfy(
   # forbidding identical rows
   AllDifferentList(x[i] for i in range(n)),  

   # forbidding identical columns
   AllDifferentList(x[:, j] for j in range(n))  
)
\end{python}\end{boxpy} 

This model involves 1 array of variables, and 3 types of constraints: \gb{Sum}, \gb{Regular}, and \gb{AllDifferentList} (which are transformed into \gb{Extension} constraints for the competition, as explained below).
Actually, depending on the chosen variant, either \gb{Sum} constraints are posted, or \gb{Regular} constraints are posted.
Note that valid rows and columns are guaranteed by the way automatas are constructed: we ensure that we have the same number of 0s and 1s, while forbidding sequences of 3 consecutive 0s or 1s.
Because \gb{AllDifferentList} is not within the perimeter of the 2023 competition (a mistake that will be fixed in 2024), such constraints have been translated into extensional forms (i.e., starred tables by calling the method \verb!to_table()!) as in:
\begin{quote}
\verb!AllDifferentList(x[i] for i in range(n)).to_table()!
\end{quote}

A series of $2*6$ instances has been selected for the competition (6 per variant).
For generating an \x3 instance (file), you can execute for example:
\begin{command}
python BinaryPuzzle.py -data=100
python BinaryPuzzle.py -data=100 -variant=regular
\end{command}

Note that when you omit to write `-variant=regular', you get the main variant.

\subsection{Calvin Puzzle}

\paragraph{Description.}

From "An Exercise for the Mind: A 10 by 10 Math Puzzle: A Pattern Recognition Game: Meditation on an Open Maze" at \href{https://chycho.blogspot.com/2014/01/an-exercise-for-mind-10-by-10-math.html}{http://www.chycho.com}.
The purpose of the game is to fill a grid of size $n \times n$ with all values ranging from 1 to $n^2$ such that:
\begin{itemize}
\item if the next number in the sequence is going to be placed vertically or horizontally, then it must be placed exactly three squares away from the previous number (there must be a two square gap between the numbers);
\item if the next number in the sequence is going to be placed diagonally, then it must be placed exactly two squares away from the previous number (there must  be a one square gap between the numbers). 
\end{itemize}

\paragraph{Data.}
Only one integer is required to specify a specific instance: the order $n$ of the grid. The values of $n$ used for generating the 2023 competition instances are: 
\begin{quote}
5, 6, 7, 8, 9, 10, 12
\end{quote}

\paragraph{Model.}
The \p3 model, in a file `CalvinPuzzle.py', used for the competition is: 

\begin{boxpy}\begin{python}
@\imp@

n = data

# x[i][j] is the value in the grid at row i and column j
x = VarArray(size=[n, n], dom=range(1, n * n + 1))

# possible neighbours
offsets = [(-3,0), (3,0), (0,-3), (0,3), (-2,-2), (-2,2), (2,-2), (2,2)]
N = [[[x[i + oi][j + oj] for (oi, oj) in offsets if 0 <= i + oi < n and 0 <= j + oj < n]
      for j in range(n)] for i in range(n)]

satisfy(
   # putting all values from 1 to n*n in the grid
   AllDifferent(x),

   # tag(symmetry-breaking)
   x[0][0] == 1
)

if not variant():
   satisfy(
      # each cell must be linked to its neighbors
      If(
         x[i][j] < n * n,
         Then=Exist(y == x[i][j] + 1 for y in N[i][j])
      ) for i in range(n) for j in range(n)
   )

elif variant("table"):

   def T(i, j):
      r = len(N[i][j]) + 1
      return [tuple(k if i == 0 else (k + 1) if i == j else ANY for i in range(r))
                for k in range(1, n * n) for j in range(1, r)] 
              + [(n * n, *[ANY] * (r - 1))]

   satisfy(
      # each cell must be linked to its neighbors
      (x[i][j], N[i][j]) in T(i, j) for i in range(n) for j in range(n)
   )
\end{python}\end{boxpy} 

This model only involves 1 array of variables $x$.
The list of variables defined as neighbors of a variable $x[i][j]$ is (computed and) given by $N[i][j]$.
Depending on the chosen variant, either \gb{Intension} and \gb{Count} (from \gb{Exist}) constraints are posted, or \gb{Extension} constraints are posted.

A series of $2*7$ instances has been selected for the competition (7 per variant).
For generating an \x3 instance (file), you can execute for example:
\begin{command}
python CalvinPuzzle.py -data=10
python CalvinPuzzle.py -data=10 -variant=table 
\end{command}

Note that when you omit to write `-variant=table', you get the main variant.

\subsection{Coloring}

\paragraph{Description.}
This is the classical graph coloring problem: given an undirected graph, one has to color the nodes of this graph such that no two adjacent nodes have the same color.

\paragraph{Data.}
As an illustration of data specifying an instance of this problem, we have:
\begin{json}
{
  "n": 450,
  "nColors": 5,
  "edges": [[0, 329], [0, 366], [0, 388], ...]
}
\end{json}

\paragraph{Model.}
The \p3 model, in a file `Coloring.py', used for the competition is: 

\begin{boxpy}\begin{python}
@\imp@

nNodes, nColors, edges = data

# x[i] is the color assigned to the ith node of the graph
x = VarArray(size=nNodes, dom=range(nColors))

satisfy(
   # two adjacent nodes must be colored differently
   [x[i] != x[j] for (i, j) in edges],

   # tag(symmetry-breaking)
   [x[i] <= i for i in range(min(nNodes, nColors))]
)
\end{python}\end{boxpy}

This model involves 1 array of variables and 1 type of constraints: \gb{Intension}.
A series of 10 instances has been selected for the competition.
For generating an \x3 instance (file), you can execute for example:
\begin{command}
python Coloring.py -data=graph1.json
\end{command}
where `graph1.json' is a data file in JSON format.

\subsection{Covering Array}

This is Problem \href{https://www.csplib.org/Problems/prob050/}{045} on CSPLib, called the Covering Array problem.

\paragraph{Description {\small (excerpt from CSPLib)}.}

A covering array $C(t,k,g,b)$ is a $k \times b$ array $A=(a_{ij})$ over $Z_g=0,1,2,\dots,g-1$ with the property that for any $t$ distinct rows $1 \leq r_1 < r_2 < \dots < r_t \leq k$, and any member $(x_1,x_2\dots,x_t)$ of $Z_g^t$ there exists at least one column $c$ such that $x_i$ equals the $(r_i,c)$-th element of $A$ for all $1 \leq i \leq t$.
Informally, any $t$ distinct rows of the covering array must encode column-wise all numbers from 0 to $g^t-1$ (repetitions being allowed).

\paragraph{Data.}
Four integers are required to specify a specific instance. Values of $(t,k,g,b)$ used for the instances in the competition are:
\begin{quote}
(3,4,2,8), (3,5,2,10), (3,6,2,12), (3,7,2,12), (3,8,2,12), (3,9,2,12), (3,10,2,12), (3,11,2,12), (4,6,2,21), (4,7,2,38), (4,8,2,42), (4,9,2,50)
\end{quote}

\paragraph{Model.}
The \p3 model, in a file `CoveringArray.py', used for the competition is:

\begin{boxpy}\begin{python}
@\imp@

t, k, g, b = data
n = factorial(k) // factorial(t) // factorial(k - t)
d = g ** t

T = {tuple(sum(pr[a] * g ** i for i, a in enumerate(reversed(co)))
      for co in combinations(range(k), t)) for pr in product(range(g), repeat=k)}

# p[i][j] is one of the position of the jth value of the ith 't'-combination
p = VarArray(size=[n, d], dom=range(b))

# v[i][j] is the jth value of the ith 't'-combination
v = VarArray(size=[n, b], dom=range(d))

satisfy(
   # all values must be present in each 't'-combination
   [AllDifferent(p[i]) for i in range(n)],

   [Channel(p[i], v[i]) for i in range(n)],

   # computing values
   [v[:, j] in T for j in range(b)]
)
\end{python}\end{boxpy}

This model involves 2 arrays of variables, and 3 types of constraints: \gb{AllDifferent}, \gb{Channel}, and \gb{Extension}.
Note that it is possible to get the covering array from the array of variables $v$.
For example, $v[0][0]$ gives the $t$ most significant bits of the first column (because the first $t$-combination is for the first $t$ lines).

A series of 12 instances has been selected for the competition.
For generating an \x3 instance (file), you can execute for example:
\begin{command}
python CoveringArray.py -data=[3,5,2,10]
\end{command}

\subsection{Dominoes}

\paragraph{Description.}

You are given a grid of size $n \times m$ containing numbers being parts of dominoes.
For example, for $n=7$ and $m=8$, the grid contains all dominoes from 0-0 to 6-6.
One has to find the position (and rotation) of each domino; see e.g., \cite{S_teaching,S_dominoes}.

\paragraph{Data.}
As an illustration of data specifying an instance of this problem, we have:
\begin{json}
{
  "grid":  [
    [0,5,2,2,5,4,6,5],
    [3,6,2,2,4,4,4,1],
    [3,6,1,2,3,4,6,1],
    [0,1,4,3,0,2,2,1],
    [3,5,3,0,3,1,5,6],
    [6,4,0,3,6,0,4,1],
    [1,6,0,0,2,5,5,5]
  ]
}
\end{json}

\paragraph{Model.}
The \p3 model, in a file `Dominoes.py', used for the competition is:

\begin{boxpy}\begin{python}
@\imp@

grid = data
nRows, nCols, nValues = len(grid), len(grid[0]), len(grid)
dominoes = [(i, j) for i in range(nValues) for j in range(i, nValues)]
indexes = range(nRows * nCols)  # indexes of cells

P = [[i * nCols + j for i in range(nRows) for j in range(nCols) if grid[i][j] == v]
       for v in range(nValues)]  # possible positions

# x[i][j] is the position in the grid of the value i of the domino i-j
x = VarArray(size=[nValues, nValues], dom=lambda i, j: indexes if i <= j else None)

# y[i][j] is the position in the grid of the value j of the domino i-j
y = VarArray(size=[nValues, nValues], dom=lambda i, j: indexes if i <= j else None)

satisfy(
   # each part of each domino in a different cell
   AllDifferent(x + y),

   # unary constraints
   [
      (
         x[i][j] in P[i],
         y[i][j] in P[j]
      ) for i, j in dominoes
   ],

   # adjacency constraints
   [
      If(
         dist != nCols,  # if not same column
         Then=both(dist == 1, x[i][j] // nCols == y[i][j] // nCols)  # then same line
      ) for i, j in dominoes if (dist := abs(x[i][j] - y[i][j]),)
   ]            
)
\end{python}\end{boxpy} 

This model involves 2 arrays of variables and 3 types of constraints: \gb{AllDifferent}, \gb{Extension} and \gb{Intension}.
A series of 12 instances has been selected for the competition.
For generating an \x3 instance (file), you can execute for example:
\begin{command}
python Dominoes.py -data=grid.json
\end{command}
where `grid.json' is a data file in JSON format.

\subsection{Fischer}

This problem has already been selected in previous XCSP competitions.
The instances selected for the 2023 competition come from a series of Fischer SMT Instances from MathSat converted to CSP by Lucas Bordeaux.
The translation is a straightforward encoding of the SMT syntax in which Boolean combinations of arithmetic constraints are decomposed into primitive constraints, using reification where appropriate.
Note that the domains have been artificially bounded, whereas in SMT theorem proving should be done over the unbounded integers.

%No \p3 model is currently available.
A series of 8 instances has been selected for the competition.

\subsection{Magic Square}

This problem has already been selected in previous XCSP competitions.
A series of 14 instances has been selected for the competition (half of the instances involving the constraint \gb{MDD}).

\subsection{Nonogram}

This problem has already been selected in previous XCSP competitions.
This series of instances is from Trieu Hung Tran (when studying, in 2019, with Berthe Y. Choueiry at the University of Nebraska-Lincoln).
He has generated these instances (involving ternary tables) from those available at \href{xcsp.org}{xcsp.org}, by using the reformulation proposed in \cite{BHHKW_slide}.
It may be the case that these instances take less space and can be solved more quickly that the original instances.

%No \p3 model is currently available.
A series of 8 instances has been selected for the competition.

\subsection{Non Transitive Dice}

\paragraph{Description.}

A set of dice is intransitive if the binary relation ``X rolls a higher number than Y more than half the time'' on its elements is not transitive.
This situation is similar to that in the game Rock, Paper, Scissors, in which each element has an advantage over one choice and a disadvantage to the other.
The problem is to exhibit such a set of dice.
See \href{https://en.wikipedia.org/wiki/Intransitive_dice}{wikipedia}.

\paragraph{Data.}
Three integers are required to specify a specific instance: the number $n$ of dices, the number $m$ of sides on each die, and the number $d$ of possible values (from $0$ to $d-1$) to be print on die sides.
Values of $(n,m,d)$ used for generating the 2023 competition instances are:
\begin{quote}
  (06,06,0), (08,08,0), (08,08,3), (10,10,0), (10,10,3), (15,15,3), \\
  (15,15,4), (20,20,3), (20,20,4), (30,30,3), (30,30,4), (40,40,0)
\end{quote}

\paragraph{Model.}
The \p3 model, in a file `NonTransitiveDice.py', following the model described by \href{http://www.hakank.org/common_cp_models/#nontransitivedice}{Hakan Kjellerstrand}, and used for the competition is: 

\begin{boxpy}\begin{python}
@\imp@

n, m, d = data # number of dice, number of sides of each die, and number of possible values
d = 2 * m if d == 0 else d  # computing the number of possible values when 0
P  = [(r1, r2) for r1 in range(m) for r2 in range(m)]

# x[i][j] is the value of the jth face of the ith die
x = VarArray(size=[n, m], dom=range(d))

# y[i] is the number of winnings of the ith die against the i+1th die (two directions)
y = VarArray(size=[n, 2], dom=range(m * m + 1))

# gap[i] is the dominance gap of the ith die
gap = VarArray(size=n, dom=range(1, m * m + 1))

# z is the maximal value on die sides
z = Var(dom=range(d))

satisfy(
   # tag(symmetry-breaking)  ordering numbers on each die 
   [Increasing(x[i]) for i in range(n)],

   # computing dominance
   [
      [y[i][0] == Sum(x[i][r1] > x[(i + 1) % n][r2] for r1, r2 in P) for i in range(n)],
      [y[i][1] == Sum(x[(i + 1) % n][r1] > x[i][r2] for r1, r2 in P) for i in range(n)]
   ],

   # computing dominance gap
   [gap[i] == y[i][0] - y[i][1] for i in range(n)],

   # computing z
   z == Maximum(x)
)
\end{python}\end{boxpy}

This model involves 3 arrays of variables, a stand-alone variable, and  4 types of constraints: \gb{Increasing}, \gb{Sum}, \gb{Intension} and \gb{Maximum}.
A series of 12 instances has been selected for the competition.
For generating an \x3 instance (file), you can execute for example:
\begin{command}
python NonTransitiveDice.py -data=[8,8,3]
\end{command}

\subsection{Peg Solitaire}

This is Problem \href{https://www.csplib.org/Problems/prob037}{037} on CSPLib.

\paragraph{Description.}

From \cite{JMMT_peg}: ``Peg Solitaire is played on a board with a number of holes. In the English version of
the game considered here, the board is in the shape of a cross with 33 holes.
Pegs are arranged on the board so that at least one hole remains. By making horizontal or vertical draughts-like moves, the pegs are gradually removed until a goal configuration is obtained.
In the classic ‘central’ Solitaire, the goal is to reverse the starting position, leaving just a single peg in the central hole.''

\paragraph{Data.}
English boards will be used (this will be acted by using the model variant `english').
We then have to decide what is the position of the initial hole, as well as the the number of moves (0 for removing all pegs but one).
Three integers are then required to specify a specific instance: the coordinates $(o_x,o_y)$ of the missing peg in the initial board, and the number $d$ of moves. The values of  $(o_x,o_y,d)$ used for generating the 2023 competition instances are: 
\begin{quote}
(0,2,0), (0,3,0), (0,4,0), (1,2,0), (1,3,0), (1,4,0), (2,0,0), (2,2,0), (2,3,0), (2,4,0), (2,6,0), (3,3,0)
\end{quote}

\paragraph{Model.}

The \p3 model, in a file `PegSolitaire.py', used for the competition is: 

\begin{boxpy}\begin{python}
@\imp@

from PegSolitaire_Generator import generate_boards, build_transitions

assert variant() in {"english", "french"} 

origin_x, origin_y, nMoves = data

init_board, final_board = generate_boards(variant(), origin_x, origin_y)
n, m = len(init_board), len(init_board[0])
transitions = build_transitions(init_board)
nTransitions = len(transitions)

v1 = sum(sum(v for v in row if v) for row in init_board)
v2 = sum(sum(v for v in row if v) for row in final_board)
horizon = v1 - v2
nMoves = horizon if nMoves == 0 or horizon < nMoves else nMoves
assert 0 < nMoves <= horizon

pairs = [(i, j) for i in range(n) for j in range(m) if init_board[i][j] is not None]

# x[i][j][t] is the value at row i and column j at time t
x = VarArray(size=[nMoves + 1, n, m], dom=lambda t,i,j: {0,1} if init_board[i][j] else None)

# y[t] is the move (transition) performed at time t
y = VarArray(size=nMoves, dom=range(nTransitions))

def unchanged(i, j, t):
   valid = [k for k, tr in enumerate(transitions) if (i, j) in (tr[0:2], tr[2:4], tr[4:6])]
   if len(valid) == 0:
      return None
   return conjunction(y[t] != k for k in valid) == (x[t][i][j] == x[t + 1][i][j])

def to0(i, j, t):
   valid = [k for k, tr in enumerate(transitions) if (i, j) in (tr[0:2], tr[2:4])]
   if len(valid) == 0:
      return None
   return disjunction(y[t] == k for k in valid) == both(x[t][i][j] == 1, x[t + 1][i][j] == 0)

def to1(i, j, t):
   valid = [k for k, tr in enumerate(transitions) if (i, j) == tr[4:6]]
   if len(valid) == 0:
      return None
   return disjunction(y[t] == k for k in valid) == both(x[t][i][j] == 0, x[t + 1][i][j] == 1)
   
satisfy(
   # setting the initial board
   x[0] == init_board,

   # setting the final board
   x[-1] == final_board,

   # setting transitions
   [
     [unchanged(i, j, t) for (i, j) in pairs for t in range(nMoves)],
     [to0(i, j, t) for (i, j) in pairs for t in range(nMoves)],
     [to1(i, j, t) for (i, j) in pairs for t in range(nMoves)]
   ]
)
\end{python}\end{boxpy} 

%Progressive relaxation produces eleven CSP instances from any single original FAPP optimization instance.

This model involves 2 array of variables and 2 types of constraints: \gb{Instantiation}, and \gb{Intension}.
The generator used to build boards is available on \x3 website.
Note how the expressions forming intensional constraints are rather complex.

A series of 12 instances has been selected for the competition.
For generating an \x3 instance (file), you can execute for example:
\begin{command}
python PegSolitaire.py -data=[3,3,0] -variant=english 
\end{command}

\subsection{Primes}

This problem has already been selected in previous XCSP competitions.
The instances selected for the 2023 competition come from a series created by Marc van Dongen: ``All instances are satisfiable. The domains of the variables consist of prime numbers and all constraints are linear equations.
The coefficients and constants in the equations are also prime numbers.
These instances are interesting because solving them using Gausian elimination is polynomial, assuming that the basic arithmetic operations have a time complexity of O(1).
In reality this assumption does not hold and the choice of prime numbers in the equations gives rise to large intermediate coefficients in the equations, making the basic operations more time consuming.
It was hoped this would allow to compare the trade-off between using a GAC approach on the original equations and the Gausian elimination approach.
The translation is a straightforward encoding of the SMT syntax in which Boolean combinations of arithmetic constraints are decomposed into primitive constraints, using reification where appropriate.
Note that the domains have been artificially bounded, whereas in SMT theorem proving should be done over the unbounded integers.''

%No \p3 model is currently available.
A series of 8 instances has been selected for the competition.

\subsection{Pythagorean Triples}

\paragraph{Description {\small (excerpt from Wikipedia)}.}
The Boolean Pythagorean triples problem is a problem from Ramsey theory about whether the positive integers can be colored red and blue so that no Pythagorean triples consist of all red or all blue members.
The Boolean Pythagorean triples problem was solved by Marijn Heule, Oliver Kullmann and Victor W. Marek in May 2016 through a computer-assisted proof.
More specifically, the problem asks if it is possible to color each of the positive integers either red or blue, so that no triple of integers $a$, $b$, $c$, satisfying $a^{2}+b^{2}=c^{2}$ are all the same color.
For example, in the Pythagorean triple 3, 4 and 5 ($3^{2}+4^{2}=5^{2}$), if 3 and 4 are colored red, then 5 must be colored blue. 
See \href{https://en.wikipedia.org/wiki/Boolean_Pythagorean_triples_problem}{wikipedia}.

\paragraph{Data.}
Only one integer is required to specify a specific instance: the limit $n$ of integers for checking the property. The values of $n$ used for generating the 2023 competition instances are:  
\begin{quote}
2000, 4000, 5000, 6000, 7000, 7500, 7824, 7825
\end{quote}

\paragraph{Model.}
The \p3 model, in a file `PythagoreanTriples.py', used for the competition is: 

\begin{boxpy}\begin{python}
@\imp@
from math import sqrt

n = data

def conflicts():
   t = []
   for i in range(1, n + 1):
      i2 = i * i
      for j in range(i + 1, n + 1):
         j2 = j * j
         s = i2 + j2
         if s > n * n:
            break
         sr = int(sqrt(s))
         if sr * sr == s:
            t.append((i, j, sr))
   return t

conflicts = conflicts()

# x[i] is 0 (resp., 1) if integer i is in part/subset 0 (resp., 1)
x = VarArray(size=n + 1, dom={0, 1})

satisfy(
   # setting an arbitrary value to integer 0
   x[0] == 0,

   # ensuring that no Pythagorean triple is present in the same part
   [NValues(x[i], x[j], x[k]) > 1 for i, j, k in conflicts]
)

\end{python}\end{boxpy} 

This model involves 1 array of variables and 2 types of constraints: \gb{Intension}, and \gb{NValues}.
A series of 8 instances has been selected for the competition.
For generating an \x3 instance (file), you can execute for example:
\begin{command}
python PythagoreanTriples.py -data=4000
\end{command}

\subsection{Slant}

\paragraph{Description {\small (from LP/CP Programming Contest 2022)}.}
From \href{https://github.com/lpcp-contest/lpcp-contest-2022/tree/main/problem-5}{LPCP'22}: ``The problem is the \href{https://www.puzzle-slant.com/}{Slant} puzzle and instances are taken/adapted from the game.
The problem is about a mission regarding islands and bridges.
The objective is to cover a map with bridges, all of them diagonals, and not necessarily connecting islands. The only requirements regard the number of bridges reaching islands, and the avoiding of cycles.''

\paragraph{Data.}
As an illustration of data specifying an instance of this problem, we have:

\begin{verbatim}
7
-1 1 -1 -1 -1 -1 1 -1 
-1 3 1 1 -1 2 -1 -1 
-1 2 -1 1 -1 3 1 1 
-1 -1 1 -1 -1 2 1 -1 
-1 -1 -1 2 -1 -1 -1 -1 
1 -1 1 1 -1 2 1 -1 
1 -1 2 2 2 3 -1 -1 
-1 1 -1 -1 -1 1 1 -1 
\end{verbatim}

\paragraph{Model.}
The \p3 model, in a file `Slant.py', used for the competition is: 

\begin{boxpy}\begin{python}
@\imp@

grid = data
n = len(grid)

DOWN_DIAG, UP_DIAG = 0, 1

# e[k][l] is 1 if the edge in the intermediate cell at coordinates (k,l) is an upward downward, 0 if a downward diagonal
e = VarArray(size=[n - 1, n - 1], dom={DOWN_DIAG, UP_DIAG})

# x[i][j] is the number of connected nodes (effective neighbors, or bridges) to node (i,j)
x = VarArray(size=[n, n], dom=lambda i, j: {grid[i][j]} if grid[i][j] != -1 else range(5))

# d[i][j] is the distance of node (i,j) from the root to the chain/tree it belongs
d = VarArray(size=[n, n], dom=range(n * n + 1))

def connections_of(i, j):
   # returns a list of tuples (k,l,a,ii,jj) where (ii,jj) is a possible node that
   # can be reached from (i,j)  by using a diagonal a on the edge (k,l)
   t = []
   if i > 0:
      if j > 0:
         t.append((i - 1, j - 1, DOWN_DIAG, i - 1, j - 1))
      if j < n - 1:
         t.append((i - 1, j, UP_DIAG, i - 1, j + 1))
   if i < n - 1:
      if j > 0:
         t.append((i, j - 1, UP_DIAG, i + 1, j - 1))
      if j < n - 1:
         t.append((i, j, DOWN_DIAG, i + 1, j + 1))
   return t

nodes = [(i, j) for i in range(n) for j in range(n)]
connections = [connections_of(i, j) for i, j in nodes]

satisfy(
   # computing the number of neighbors
   [x[i][j] == Sum(e[k][l] == a for (k,l,a,_,_) in connections_of(i, j)) for i, j in nodes],

   # isolated nodes are roots (are at distance 0)
   [If(x[i][j] == 0, Then=d[i][j] == 0) for i, j in nodes],

   # nodes with at least 2 neighbors cannot be roots (be at distance 0)
   [If(x[i][j] > 1, Then=d[i][j] != 0) for i, j in nodes],

   # nodes with at least 2 neighbors have exactly 1 parent (node with the same distance - 1)
   [If(x[i][j] > 1, Then=Count(both(e[k][l] == a,  d[i][j] == d[ii][jj] + 1)
     for (k, l, a, ii, jj) in connections_of(i, j)) == 1) for i, j in nodes],

   # the distance between any two neighbors is always 1
   [If(e[k][l] == a, Then=abs(d[i][j] - d[ii][jj]) == 1) for i, j in nodes
     for (k, l, a, ii, jj) in connections_of(i, j)]
)
\end{python}\end{boxpy}

This problem involves 3 array of variables and 3 type of constraints: \gb{Sum}, \gb{Intension} and \gb{Count}.
A series of 10 instances (with data coming from LPCP'22) has been selected for the competition.
For generating an \x3 instance (file), you can execute for example:
\begin{command}
python Slant.py -data=instance01.in -parser=Slant_Parser.py
\end{command}
where `instance01.in' is a data file and `Slant\_Parser.py' is a parser (i.e., a Python file allowing us to load data that are not directly given in JSON format).
Note that for saving data in JSON files, you can add the option `-export' (or `-dataexport').

\subsection{Soccer}

\paragraph{Description.}

Soccer computational problems have been studied in \cite{DAD_soccer}, and several instances have been submitted to Minizinc challenges 2018 and 2020.
This is related to \href{http://www.sabiofutbol.com/}{SABIO} \cite{DDa_sabio}, an interactive platform that can be used to represent several soccer computational problems using CP, and notably, 
``position in ranking'', which let users to impose constraints about the positions of the teams at the end of a tournament.

\paragraph{Data.}
As an illustration of data specifying an instance of the ``position in ranking'' problem, we have:

\begin{json}
{
  "games": [[1, 0], [2, 0], [2, 1], ..., [23, 22]],
  "initial_points": [21, 21, 21, ...16],
  "positions": [[0, 21], [1, 19], [2, 17], ..., [13, 3]]
}
\end{json}

\paragraph{Model.}
The \p3 model, in a file `Soccer.py', based on the model submitted by Robinson Duque, Alejandro Arbelaez, and Juan Francisco Díaz to the Minizinc challenges 2018 and 2020, and used for the competition is: 

\begin{boxpy}\begin{python}
@\imp@

games, iPoints, positions = data
nGames, nTeams, nPositions = len(games), len(iPoints), len(positions)

pt = [0, 1, 3]

lb_score = min(iPoints[i] + sum(min(pt) for j in range(nGames) if i in games[j])
     for i in range(nTeams))
ub_score = max(iPoints[i] + sum(max(pt) for j in range(nGames) if i in games[j])
     for i in range(nTeams))

# points[j] are the points for the two teams (indexes 0 and 1) of the jth game
points = VarArray(size=[nGames, 2], dom=pt)

# score[i] is the final score of the ith team
score = VarArray(size=nTeams, dom=range(lb_score, ub_score + 1))

# fp[i] is the final position of the ith team
fp = VarArray(size=nTeams, dom=range(1, nTeams + 1))

# bp[i] is the best possible position of the ith team
bp = VarArray(size=nTeams, dom=range(1, nTeams + 1))

# wp[i] is the worst possible position of the ith team
wp = VarArray(size=nTeams, dom=range(1, nTeams + 1))

satisfy(
   # assigning rights points for each game
   [(points[j][0], points[j][1]) in {(0, 3), (1, 1), (3, 0)} for j in range(nGames)],

   # computing final points
   [score[i] - Sum(points[j][0 if i == games[j][0] else 1] for j in range(nGames) if i in games[j]) == iPoints[i] for i in range(nTeams)],

   # computing worst positions (the number of teams with greater total points)
   [wp[i] == Sum(score[j] >= score[i] for j in range(nTeams)) for i in range(nTeams)],

   # computing best positions (from worst positions and number of teams with equal points)
   [bp[i] == wp[i] - Sum(score[j] == score[i] for j in range(nTeams) if i != j)
     for i in range(nTeams)],

   # bounding final positions
   (
      [fp[i] >= bp[i] for i in range(nTeams)],
      [fp[i] <= wp[i] for i in range(nTeams)]
   ),

   # ensuring different positions
   AllDifferent(fp),

   # applying rules from specified positions
   (
      [fp[i] == p for i, p in positions],
      [Sum(score[j] > score[i] for j in range(nTeams) if j != i) < p for i, p in positions],
      [Sum(score[j] < score[i] for j in range(nTeams) if j != i) <= nTeams + 1 - p
        for i, p in positions]
   )
)
\end{python}\end{boxpy}

This problem involves 5 array of variables and 4 type of constraints: \gb{Extension}, \gb{Sum}, \gb{Intension} and \gb{AllDiferent}.
A series of 10 instances (data having courteously been provided by R. Duque) has been selected for the competition.
For generating an \x3 instance (file), you can execute for example:
\begin{command}
python Soccer.py -data=30-20-24-1.json 
\end{command}
where `30-20-24-1.json' is a data file in JSON format.

\subsection{Square Packing Suite}

\paragraph{Description.}
The Square packing problem involves packing all squares with sizes $1 \times 1$ to $n \times n$ into an enclosing container \cite{MP_optimal,SO_search}.

\paragraph{Data.}
Only one integer is required to specify a specific instance: the order $n$ (i.e., the largest square to be packed). The values of $n$ used for generating the 2023 competition instances are: 
\begin{quote}
15, 18, 20, 21, 22, 23, 24, 25, 26, 27 
\end{quote}

\paragraph{Model.}
The \p3 model, in a file `SquarePackingSuite.py', used for the competition is: 

\begin{boxpy}\begin{python}
@\imp@

n = data
assert 6 <= n <= 27  # for data (containers) below, as given in papers mentioned above

containers = [[9, 11], [7, 22], [14, 15], [15, 20],  [15, 27],  [19, 27],  [23, 29],  [22, 38],  [23, 45],  [23, 55],  [27, 56],  [39, 46],  [31, 69],  [47, 53],  [34, 85],  [38, 88],  [39, 98],  [64, 68],  [56, 88],  [43, 129],  [70, 89],  [47, 148]]  # indices from 6 to 27

# initial reduction as indicated in the CP'08 paper
t = [[], [1, 2], [2, 3], [2]] + [[3]] * 4 + [[4]] * 3 + [[5]] * 6 + [[6]] * 4
    + [[7]] * 8 + [[8]] * 5 + [[9]] * 11 + [[10]]

width, height = containers[n - 6]

# x[i] is the x-coordinate where is put the ith rectangle
x = VarArray(size=n, dom=lambda i: range(width - i))

# y[i] is the y-coordinate where is put the ith rectangle
y = VarArray(size=n, dom=lambda i: range(height - i))

satisfy(
   # no overlap on boxes
   NoOverlap(origins=[(x[i],y[i]) for i in range(n)], lengths=[(i+1,i+1) for i in range(n)]),

   # tag(redundant-constraints)
   [
       Cumulative(
          Task(origin=x[i], length=i + 1, height=i + 1) for i in range(n)
       ) <= height,

       Cumulative(
          Task(origin=y[i], length=i + 1, height=i + 1) for i in range(n)
       ) <= width
   ],

   # tag(symmetry-breaking)
   (
       [x[-1] <= (width - n) // 2, y[-1] <= (height - n) // 2],
       [x[i] != v for i in range(n) for v in t[i]],
       [y[i] != v for i in range(n) for v in t[i]]
   )
)
\end{python}\end{boxpy} 

This model involves 2 array of variables, and 3 type of constraints: \gb{NoOverlap}, \gb{Cumulative}, and \gb{Intension}.
A series of $10$ instances has been selected for the competition.
For generating an \x3 instance (file), you can execute for example:
\begin{command}
python SqurePackingSuite.py -data=20
\end{command}

\subsection{Word Design (for DNA Computing on Surfaces)}

This is Problem \href{https://www.csplib.org/Problems/prob033/}{033} on CSPLib.

\paragraph{Description {\small (excerpt from CSPLib)}.}

This problem (proposed by M. Van Dongen) on CSPLib has its roots in Bioinformatics and Coding Theory.
It is to find as large as possible a set $S$ of strings (words) of length 8 over the alphabet $W = \{ A,C,G,T \}$ with the following properties:
\begin{itemize}
\item each word in S contains 4 occurrences of symbols from $\{ C,G \}$,
\item each pair of distinct words in $S$ differ in at least 4 positions, 
\item each pair of words $x$ and $y$ in $S$ (where $x$ and $y$ may be identical) are such that $x^R$ and $y^C$ differ in at least 4 positions;
  Here, $(x_1,\dots,x_8)^R = (x_8,\dots,x_1)$ is the reverse of $(x_1\dots,x_8)$ and $(y_1\dots,y_8)^C$ is the Watson-Crick complement of $(y_1\dots,y_8)$, i.e. the word where each A is replaced by a T and vice versa and each C is replaced by a G and vice versa.
\end{itemize}

\paragraph{Data.}
A first invariant JSON file, called `words.json', indicates the possible words (each word has 4 symbols from $\{1,2\} = \{C,G \}$) and is such that its reverse and Watson-Crick complement differ in at least 4 positions):
\begin{json}
{
  "words": [
    [0,0,0,0,1,1,1,1],
    [0,0,0,0,1,1,1,2],
    [0,0,0,0,1,1,2,1],
    [0,0,0,0,1,1,2,2],
    ...
  ]
}
\end{json}

A second invariant JSON, called `mdd.json', indicates the transitions of a MDD that can be used to enforce the restrictions on pairs of words (it will be used for a second model): 
\begin{json}
{
  "transitions":[
    ["root",0,"n3"],
    ["root",1,"n286276"],
    ["root",2,"n430777"],
    ...
  ]
}
\end{json}

Finally, CSP instances can be generated by setting the size $n$ of the set $S$.  
The values of $n$ used for generating the 2023 competition instances are: 
\begin{quote}
5, 15, 25, 35, 45, 55, 65, 75, 85, 100
\end{quote}

\paragraph{Model.}
A first \p3 model, in a file `WordDesign1.py', used for the competition is: 

\begin{boxpy}\begin{python}
@\imp@

words, n = data  

# x[i][k] is the kth letter (0-A, 1-C, 2-G, 3-T) of the ith word
x = VarArray(size=[n, 8], dom=range(4))

# y[i][k] is the kth letter of the Watson-Crick complement of the ith word (in x)
y = VarArray(size=[n, 8], dom=range(4))

satisfy(
   # computing the Watson-Crick complement of words
   [x[i][k] + y[i][k] == 3 for i in range(n) for k in range(8)],

   # each word must be well formed
   [x[i] in words for i in range(n)],

   # ordering words  tag(symmetry-breaking)
   LexIncreasing(x, strict=True),

   # each pair of distinct words differ in at least 4 positions
   [Sum(x[i][k] != x[j][k] for k in range(8)) >= 4 for i, j in combinations(n, 2)],

   # each pair of distinct words are such that the reverse of the former and the Watson-Crick complement of the latter differ in at least 4 positions
   [Sum(x[i][7 - k] != y[j][k] for k in range(8)) >= 4 for i in range(n) for j in range(n)
     if i != j]
)
\end{python}\end{boxpy} 

This first model involves 2 arrays of variables and 4 types of constraints: \gb{Intension}, \gb{Extension}, \gb{LexIncreasing} and \gb{Sum}.
A series of 10 instances has been selected for the competition.
For generating an \x3 instance (file), you can execute for example:
\begin{command}
python WordDesign1.py -data=[words.json,n=15]
\end{command}
Note how we can append a specific parameter to the data coming from a JSON file.

\bigskip
A second \p3 model, in a file `WordDesign2.py', used for the competition is: 

\begin{boxpy}\begin{python}
@\imp@

words, transitions, n = data  
M = MDD(transitions)

# x[i][k] is the kth letter (0-A, 1-C, 2-G, 3-T) of the ith word
x = VarArray(size=[n, 8], dom=range(4))

satisfy(
   # each word must be well-formed
   [x[i] in words for i in range(n)],

   # ordering words  tag(symmetry-breaking)
   LexIncreasing(x, strict=True),

   # ensuring the validity of any pair of words
   [x[i] + x[j] in M for i, j in combinations(n, 2)]
)
\end{python}\end{boxpy} 

This second model involves 1 array of variables and 3 types of constraints: \gb{Extension}, \gb{LexIncreasing} and \gb{MDD}.
A series of 10 instances has been selected for the competition.
For generating an \x3 instance (file), you can execute for example:
\begin{command}
python WordDesign2.py -data=[words.json,mdd.json,n=15]
\end{command}
Note how we can append a specific parameter to the data coming from two JSON files.

\section{COP}

\subsection{Aircraft Assembly Line}

\paragraph{Description.}

This problem has been proposed by Stéphanie Roussel from ONERA (Toulouse), and comes from an aircraft manufacturer.
The objective is to schedule tasks on an aircraft assembly line in order to minimize the overall number of operators required on the line.
The schedule must satisfy several operational constraints, the main ones being:
\begin{itemize}
\item tasks are assigned on a unique workstation (on which specific machines are available);
\item the takt-time, i.e., the duration during which the aircraft stays on each workstation, must be respected;
\item capacity of aircraft zones in which operators perform the tasks must never be exceeded;
\item zones can be neutralized by some tasks, i.e., it is not possible to work in those zones during the tasks execution.
\end{itemize}

Note that similar problems have been studied in \cite{RCP23}, where the authors are interested in the design of assembly lines (with similar instances).

\paragraph{Data.}
As an illustration of data specifying an instance of this problem, we have:

\begin{json}
{
  "takt": 1440,
  "nTasks": 199,
  "nMachines": 5,
  "nAreas": 48,
  "areasCapacities": [1,1,..,1],
  "tasksPerMachine": [
    [49,50,51,52,53],
    ...,
    [197]
  ],
  "nMaxOpsPerStation": [10,10,10,10],
  "neutralizedAreas": [
    [12,15,24,28,31,37,39,46],
    ...,
    []
  ],
  "operators": [0,1,...,1],
  "tasksPerAreas": [
    [2,5,11,13],
    ...,
    [6,7,165,166,167]
  ],
  "usedAreas":[
    [0,0,...,0],
    ...,
    [0,0,...,0]
  ],
  "durations": [0,109,...,470],
  "precedences": [
    [8,9],
    ...,
    [198,190]
  ],
  "nStations": 4,
  "machines":[
    [0,0,0,0,0],
    ...,
    [0,0,1,1,1]
  ]
}
\end{json}

\paragraph{Model.}
\bigskip
The \p3 model, in a file `AircraftAssemblyLine.py', used for the competition is:

\begin{boxpy}\begin{python}
@\imp@

takt, areas, stations, tasks, tasksPerMachine, precedences = data
nAreas, nStations, nTasks = len(areas), len(stations), len(tasks)
nMachines = len(tasksPerMachine)

areaCapacities, areaTasks = zip(*areas)  # nb of operators who can work, and tasks per area
stationMachines, stationMaxOperators = zip(*stations)
durations, operators, usedAreaRooms, neutralizedAreas = zip(*tasks)
usedAreas = [set(j for j in range(nAreas) if usedAreaRooms[i][j] > 0) for i in range(nTasks)]

def station_of_task(i):
   r = next((j for j in range(nMachines) if i in tasksPerMachine[j]), -1)
   return -1 if r == -1 else next(j for j in range(nStations) if stationMachines[j][r] == 1)

stationOfTasks = [station_of_task(i) for i in range(nTasks)]  # -1 if can be everywhere

# x[i] is the starting time of the ith task
x = VarArray(size=nTasks, dom=range(takt * nStations + 1))

# z[j] is the number of operators at the jth station
z = VarArray(size=nStations, dom=lambda i: range(stationMaxOperators[i] + 1))

satisfy(
   # respecting the final deadline
   [x[i] + durations[i] <= takt * nStations for i in range(nTasks)],

   # ensuring that tasks start and finish in the same station
   [x[i] // takt == (x[i] + max(0, durations[i] - 1)) // takt
      for i in range(nTasks) if durations[i] != 0],

   # ensuring that tasks are put on the right stations (wrt needed machines)
   [x[i] // takt == stationOfTasks[i] for i in range(nTasks) if stationOfTasks[i] != -1],

   # respecting precedence relations
   [x[i] + durations[i] <= x[j] for (i, j) in precedences],

   # respecting limit capacities of areas
   [
      Cumulative(
         Task(origin=x[t], length=durations[t], height=usedAreaRooms[t][i])
            for t in areaTasks[i]
      ) <= areaCapacities[i] for i in range(nAreas) if len(areaTasks[i]) > 1
   ],

   # computing/restricting the number of operators at each station
   [
      Cumulative(
         Task(origin=x[t], length=durations[t], height=operators[t] * (x[t] // takt == j))
            for t in range(nTasks)
      ) <= z[j] for j in range(nStations)
   ],
    
   # no overlapping between some tasks 
   [NoOverlap(tasks=[(x[i], durations[i]), (x[j], durations[j])])
      for i in range(nTasks) for j in range(nTasks)
         if i != j and len(usedAreas[i].intersection(neutralizedAreas[j])) > 0],

   # avoiding tasks using the same machine to overlap
   [NoOverlap(tasks=[(x[j], durations[j]) for j in tasksPerMachine[i]])
      for i in range(nMachines) if len(tasksPerMachine[i]) > 1]
)

minimize(
   # minimizing the number of operators
   Sum(z)
)
\end{python}\end{boxpy} 

This involves 2 arrays of variables and 3 types of constraints: \gb{Intension}, \gb{Cumulative} and \gb{NoOverlap}.
A series of 20 instances has been selected from data files generated by Stéphanie Roussel from ONERA (Toulouse).

For generating an \x3 instance (file), you can execute for example:
\begin{command}
python AircraftAssemblyLine.py -data=1-178.json -parser=Aircraft_Converter.py
\end{command}

where `1-178.json' is a data file in JSON format, and `Aircraft\_Converter.py' is a converter tool allowing us to pass from one JSON format to another that is easier to handle (with respect to the model).
See how the initial 15 data fields have been reduced to 6 data fields only. 
Note that for saving data in JSON files, you can add the option `-export' (or `-dataexport').

\subsection{Beer Jugs}

\paragraph{Description {\small (from LP/CP Programming Contest 2022)}.}
From \href{https://github.com/lpcp-contest/lpcp-contest-2022/tree/main/problem-2}{LPCP'22}: ``Blue Meth, Blue Sky, Heisenberg Blue... movie directors often like to adapt scientific (or maybe scientif-ish) concepts in their creations.
You are asked to help some of those fellas, who were impressed by the water jugs riddle in Die Hard 3.
They want to shoot a similar scene, but in a pub, using beer instead of water (obviously, no beer will be wasted, and a drop essentially means that the main actors drink all the beer from a jug).
In order to do the scene as much comical as possible, they ask for the longest sequence of non-repeating configurations that can be achieved using two jugs.
Given the capacities of the two jugs, A and B, the possible actions are the following:
\begin{itemize}
    \item drop\_a, to empty the first jug;
    \item drop\_b, to empty the second jub;
    \item fill\_a, to fill the first jug;
    \item fill\_b, to fill the second jug;
    \item a\_to\_b, to pour the second jug with the content of the first jug (either until the second jug is full, or until the first jug is empty);
    \item b\_to\_a, to pour the first jug with the content of the second jug (either until the first jug is full, or until the second jug is empty).
\end{itemize}
Note that they don't care if the final configuration can be achieved with a shorter sequence... no one will note, they'll all be drunk for a while anyway.''

\paragraph{Data.}
Two integers are required to specify a specific instance: the capacities $A$ and $B$ of the two jugs. The values of $(A,B)$ used for generating the 2023 competition instances are (some of the LPCP'22 contest): 
\begin{quote}
(3,10), (9,10), (7,12), (11,12), (11,14), (11,16), (13,16), (15,16)
\end{quote}

\paragraph{Model.}
The \p3 model, in a file `BeerJugs.py', used for the competition is: 

\begin{boxpy}\begin{python}
@\imp@

A, B = data
MAX = 70

STOP, FILL_A, FILL_B, DROP_A, DROP_B, A_TO_B, B_TO_A = Actions = range(-1, 6)

def execute(q1, q2, action):
   if action == STOP:
      return -1, -1
   if action == FILL_A:
      return (A, q2) if q1 != A else None
   if action == FILL_B:
      return (q1, B) if q2 != B else None
   if action == DROP_A:
      return (0, q2) if q1 > 0 else None
   if action == DROP_B:
      return (q1, 0) if q2 > 0 else None
   if action == A_TO_B:
      pour = min(q1, B - q2)
      return (q1 - pour, q2 + pour) if pour > 0 else None
   if action == B_TO_A:
      pour = min(A - q1, q2)
      return (q1 + pour, q2 - pour) if pour > 0 else None

valid = [(q1, q2, a) for q1 in range(A + 1) for q2 in range(B + 1) for a in Actions
          if execute(q1, q2, a)]
T = [(-1, -1, -1, -1, -1)] + [(q1, q2, a, *execute(q1, q2, a)) for q1, q2, a in valid]

# x[t][i] is the quantity in the ith jug (i is equal to 0 for A and 1 for B) at time t
x = VarArray(size=[MAX+1,2], dom=lambda i,j: {0} if i==0 else range(-1,(A if j==0 else B)+1))

# y[t] is the action taken at time t (to t+1)
y = VarArray(size=MAX, dom=Actions)

# z is the time when the process is stopped
z = Var(range(MAX))

satisfy(
   # ensuring that the same state is never encountered several times
   If(
      s1[0] != -1,
      Then=either(s1[0] != s2[0], s1[1] != s2[1])
   ) for s1, s2 in combinations(x, 2),

   # computing the consequences of each action
   [(x[t][0], x[t][1], y[t], x[t + 1][0], x[t + 1][1]) in T for t in range(MAX)],

   # ensuring a stable state (-1, -1) when the process is finished
   [(t < z) == (y[t] != STOP) for t in range(MAX)]
)

maximize(
   # maximizing the length of the sequence of actions
   z
)
\end{python}\end{boxpy}

This problem involves 2 arrays of variables, a stand-alone variable and 2 type of constraints: \gb{Intension} and \gb{Extension}.
A series of 8 instances (with data coming from LPCP'22) has been selected for the competition.
For generating an \x3 instance (file), you can execute for example:
\begin{command}
python BeerJugs.py -data=[9,10]
\end{command}

\subsection{Benzenoide}

\paragraph{Description.}

From \cite{CHPTV_how}: ``The benzenoid generation problem is defined as follows: given a set of structural properties P, generate all the benzenoids which satisfy each property of P.
For instance, these structural properties may deal with the number of carbons, the number of hexagons or a particular structure for the hexagon graph.''
Here, we are interested in generating benzenoids with $n$ hexagons (including benzenoids with `holes').

%\cite{VPTHC_benzai,CHPTV_how}.

\paragraph{Data.}
Only one integer is required to specify a specific instance: the order $n$ of the coronenoide. The values of $n$ used for generating the 2023 competition instances are: 
\begin{quote}
6, 7, 8, 9, 10, 11, 12, 13, 14, 15
\end{quote}

\paragraph{Model.}
The \p3 model, in a file `Benzenoide.py', used for the competition is: 

\begin{boxpy}\begin{python}
@\imp@

n = data  # order of the coronenoide
w = 2 * n - 1  # maximal width
widths = [w - abs(n - i - 1) for i in range(w)]

symmetries = [sym.apply_on(n) for sym in TypeHexagonSymmetry]  

def valid(*t):
    return [(i, j) for i, j in t if 0 <= i < w and 0 <= j < widths[i]]

neighbors = [[valid(
    (i, j - 1), (i, j + 1),
    (i - 1, j - (1 if i < n else 0)), (i - 1, j + (0 if i < n else 1)),
    (i + 1, j - (1 if i >= n - 1 else 0)), (i + 1, j + (0 if i >= n - 1 else 1)))
      for j in range(widths[i])] for i in range(w)]

def T1(i, j):
    r = len(neighbors[i][j])
    return [(0, 0, *[ANY] * r),  (1, 1, *[ANY] * r)] + 
        [(2, 1, *[1 if j == i else ANY for j in range(r)]) for i in range(r)] + 
        [(v, 1, *[v - 1 if j == i else {0}.union(range(v - 1, n + 1)) for j in range(r)])
          for v in range(3, n + 1) for i in range(r)]

T2 = [(1,1,1,1,1,1,1)] + [(ANY, *[0 if j == i else ANY for j in range(6)]) for i in range(6)]

# x[i][j] is 1 iff the hexagon at row i and column j is selected
x = VarArray(size=[w, w], dom=lambda i, j: {0, 1} if j < widths[i] else None)

# y[i][j] is the distance (+1) wrt the root of the connected tree
y = VarArray(size=[w, w], dom=lambda i, j: range(n + 1) if j < widths[i] else None)

satisfy(
   # only one root
   Count(y, value=1) == 1,

   # ensuring connectedness
   [(y[i][j], x[i][j], y[neighbors[i][j]]) in T1(i, j)
      for i in range(w) for j in range(widths[i])],

   # exactly n hexagons
   Sum(x) == n,

   # ensuring no holes
   [(x[i][j], x[neighbors[i][j]]) in T2 for i in range(w)
      for j in range(widths[i]) if len(neighbors[i][j]) == 6],

   # tag(symmetry-breaking)
   [
      [LexDecreasing(x, [x[row] for row in sym]) for sym in symmetries],
      [Precedence(y, values=(1, v)) for v in range(2, n + 1)]
   ]
)

minimize(
   Sum(x[i][j] * ((n-i) * w + (n-j)) for i in range(w) for j in range(w) if j < widths[i])
)
\end{python}\end{boxpy} 

This model involves 2 arrays of variables, and 5 types of constraints: \gb{Count}, \gb{Extension}, \gb{Sum}, \gb{LexDecreasing} and \gb{Precedence}.
A series of 10 instances has been selected for the competition.
For generating an \x3 instance (file), you can execute for example:
\begin{command}
python Benzenoide.py -data=10
\end{command}

\subsection{Carpet Cutting}

\paragraph{Description.}

From \cite{SSV_optimal}: ``The carpet cutting problem is a two-dimensional cutting and packing problem in which carpet shapes (also called items or objects) are cut from a rectangular carpet roll with a fixed roll width and a sufficiently long roll length.
The goal is to find a non-overlapping placement of all carpet shapes on the carpet roll, so that the waste is minimized or in other words the utilization of used carpet material is maximized while meeting all additional constraints.
In our case the objective is to minimize the carpet roll length.''
  
\paragraph{Data.}
Twenty instances have been selected in several Minizinc challenges (in 2011, 2012, 2016 and 2021); see \href{https://github.com/MiniZinc/minizinc-benchmarks/tree/master/carpet-cutting}{minizinc-benchmarks}.
The structure of a data file in JSON format is as follows:

\begin{json}
{
  "roll_wid": 315,
  "max_roll_len": 20000,
  "roomCarpets": [
    {
      "rectangleIds": [0],
      "possibleRotations": [0, 1],
      "maxLength": 181,
      "mawWidth": 131
    }, ...
  ],
  "rectangles": [
    {
      "length": 181,
      "width": 131,
      "xOffsets": [0, 0, -1, -1],
      "yOffsets": [0, 0, -1, -1]
    }, ...
  ],
  "stairCarpets": []
}
\end{json}

\paragraph{Model.}
The \p3 model, in a file `CarpetCutting.py', based on the \href{https://github.com/MiniZinc/minizinc-benchmarks/blob/master/carpet-cutting/cc_base.mzn}{Minizinc model}, used for the competition is: 

\begin{boxpy}\begin{python}
@\imp@

rollWidth, maxRollLength, roomCarpets, rectangles, stairCarpets = data
roomRectangles, possibleRotations, maxLengths, maxWidths = zip(*roomCarpets)
rectLengths, rectWidths, xOffsets, yOffsets = zip(*rectangles)
stairLengths, stairWidths, nCoveredSteps, minCutSteps, maxCuts = zip(*stairCarpets)
   if len(stairCarpets) > 0 else ([], [], [], [], [])
nRoomCarpets, nRectangles, nStairCarpets = len(roomCarpets),len(rectangles),len(stairCarpets)

RC, SC = range(nRoomCarpets), range(nStairCarpets)

R0, R90, R180, R270 = Rotations = range(4)  # 0 - 0, 1 - 90, 2 - 180, and 3 - 270

stairOffsets = [sum(nCoveredSteps[:i]) for i in SC]
stairRanges = [range(sco, sco + nCoveredSteps[i]) for i, sco in enumerate(stairOffsets)]

nSteps = sum(nCoveredSteps)
stepLengths =[stairLengths[i]//nCoveredSteps[i] for i in SC for _ in range(nCoveredSteps[i])]
stepWidths = [stairWidths[i] for i in SC for _ in range(nCoveredSteps[i])]

totalArea = sum(l * w for (l, w, _, _) in rectangles) + sum(stairLengths[i] * stairWidths[i] for i in SC)
minRollLength = (totalArea // rollWidth) + (1 if totalArea % rollWidth > 0 else 0)
totalLength = sum(max(maxLengths[i], maxWidths[i]) for i in RC) + sum(stairLengths)
maxRollLength = min(maxRollLength, totalLength)

rectSizes = range(min(min(l, w) for (l, w, _, _) in rectangles), max(max(l, w)
   for (l, w, _, _) in rectangles) + 1)

nexts = [next(j for j in RC if i in roomRectangles[j]) for i in range(nRectangles)]

# z is the carpet roll length
z = Var(dom=range(minRollLength, maxRollLength + 1))

# x[i] is the x-coordinate of the ith room carpet
x = VarArray(size=nRoomCarpets, dom=range(maxRollLength + 1))

# y[i] is the y-coordinate of the ith room carpet
y = VarArray(size=nRoomCarpets, dom=range(rollWidth + 1))

# r[i] is the rotation of the ith room carpet
r = VarArray(size=nRoomCarpets, dom=Rotations)

# r0or180[is] is 1 if the rotation of the ith room carpet is 0 or 180 degrees
r0or180 = VarArray(size=nRoomCarpets, dom={0, 1})

# r0or90[is] is 1 if the rotation of the ith room carpet is 0 or 90 degrees
r0or90 = VarArray(size=nRoomCarpets, dom={0, 1})

# xr[j] is the x-coordinate of the jth rectangle
xr = VarArray(size=nRectangles, dom=range(maxRollLength + 1))

# yr[j] is the y-coordinate of the jth rectangle
yr = VarArray(size=nRectangles, dom=range(rollWidth + 1))

# lr[j] is the length of the jth rectangle (considering a possible rotation)
lr = VarArray(size=nRectangles, dom=rectSizes)

# wr[j] is the width of the jth rectangle (considering a possible rotation)
wr = VarArray(size=nRectangles, dom=rectSizes)

if nSteps > 0:
   # xs[k] is the x-coordinate of the kth step of the stair
   xs = VarArray(size=nSteps, dom=range(maxRollLength + 1))

   # ys[k] is the y-coordinate of the kth step of the stair
   ys = VarArray(size=nSteps, dom=range(rollWidth + 1))

   # lp[i][j] is 1 if the jth covered step by the ith stair carpet is the last step of a part (of the partition of the stair carpet)
   lp = VarArray(size=[nStairCarpets, nCoveredSteps], dom={0, 1})

   ls, ws = stepLengths, stepWidths

else:
   xs = ys = lp = ls = ws = []

X, Y, L, W = xr + xs, yr + ys, lr + ls, wr + ws


satisfy(
   # computing lengths and widths of rectangles
   [
      (
         lr[i] == rectWidths[i] + (rectLengths[i] - rectWidths[i]) * r0or180[nexts[i]],
         wr[i] == rectLengths[i] + (rectWidths[i] - rectLengths[i]) * r0or180[nexts[i]]
      ) for i in range(nRectangles)
   ],

   # enforcing room carpets to stay within limits
   [
      (
         x[i] + maxWidths[i] + (maxLengths[i] - maxWidths[i]) * r0or180[i] <= z,
         y[i] + maxLengths[i] + (maxWidths[i] - maxLengths[i]) * r0or180[i] <= rollWidth
      ) for i in RC
   ],

   # enforcing rectangles (of room carpets) to stay within limits
   [
      (
         xr[j] + lr[j] <= z,
         yr[j] + wr[j] <= rollWidth
      ) for i in RC for j in roomRectangles[i]
   ],

   # handling possible rotations of room carpets
   [
      (
         r[i] in possibleRotations[i],
         r0or90[i] == (r[i] in {R0, R90}),
         r0or180[i] == (r[i] in {R0, R180})
      ) for i in RC
   ],

   # computing the coordinates of the rectangles in room carpets
   [
      (
         xr[j] == x[i] + xOffsets[j][r[i]],
         yr[j] == y[i] + yOffsets[j][r[i]]
      ) for i in RC for j in roomRectangles[i]
   ],

   # enforcing stair steps to stay within limits
   [
      (
         xs[j] + stepLengths[j] <= z,
         ys[j] + stepWidths[j] <= rollWidth
      ) for i in SC for j in stairRanges[i]
   ],

   # breaking symmetries between steps of a stair carpet  tag(symmetry-breaking)
   [
      (
         ys[j] <= ys[j + 1],
         If(ys[j] >= ys[j + 1], Then=xs[j] + stepLengths[j] <= xs[j + 1])
      ) for i in SC for j in stairRanges[i] if j + 1 in stairRanges[i]
   ],

   # computing the last steps in the parts of the partitions of the stair carpets
   [
      [lp[i][nCoveredSteps[i] - 1] == 1 for i in SC],

      [
         lp[i][j] == either(
            ys[k] < ys[k + 1],
            xs[k] + stepLengths[j] < xs[k + 1]
         ) for i, offset in enumerate(stairOffsets) for j in range(nCoveredSteps[i] - 1)
             if [k := j + offset]
      ],

      [
         If(
            xs[k] + 2 * stepLengths[k] > z,
            Then=lp[i][j]
         ) for i, offset in enumerate(stairOffsets) for j in range(nCoveredSteps[i])
             if [k := j + offset]
      ],

      [Sum(lp[i][:nCoveredSteps[i]]) <= maxCuts[i] + 1 for i in SC],

      [lp[i][j] == 0 for i in SC if minCutSteps[i] > 1 for j in range(minCutSteps[i] - 1)],

      [
         If(
            lp[i][j],
            Then=lp[i][k] == 0
         ) for i in SC if minCutSteps[i] > 1 for j in range(minCutSteps[i]-1,nCoveredSteps[i])
             for k in range(j - minCutSteps[i] + 1, j)
      ]
   ],

   # respecting roll length
   Cumulative(origins=X, lengths=L, heights=W) <= rollWidth,

   # respecting roll width
   Cumulative(origins=Y, lengths=W, heights=L) <= z,

   # non-overlapping
   NoOverlap(origins=zip(X, Y), lengths=zip(L, W))
)

minimize(
   # minimizing the carpet roll length
   z
)
\end{python}\end{boxpy} 

This model involves many arrays of variables, and 5 types of constraints: \gb{Intension}, \gb{Extension}, \gb{Cumulative}, \gb{NoOverlap} and \gb{Element}. 
A series of 20 instances has been selected for the competition.
For generating an \x3 instance (file), you can execute for example:
\begin{command}
python CarpetCutting.py -data=test01.json
\end{command}
where `test01.json' is a data file in JSON format.

\subsection{Generalised Balanced Academic Curriculum Problem (GBACP)}

This is Problem \href{https://www.csplib.org/Problems/prob064}{064} on CSPLib.

\paragraph{Description.}

From CSPLib: ``This is a generalisation of the Balanced Academic Curriculum Problem (BACP) proposed by Marco Chiarandini, Luca Di Gaspero, Stefano Gualandi, and Andrea Schaerf at University of Udine.
With respect to BACP, GBACP adds:
\begin{itemize}
\item several curricula that can share courses,
\item soft constraints, in particular for teacher preferences for not teaching during some terms.
\end{itemize}
The detailed description, data, best results, and a solution validator can be found at \href{https://opthub.uniud.it/problem/timetabling/gbac}{opthub.uniud.it}''.
See also \cite{GS_hybrid}.

\paragraph{Data.}

Ten historical instances come from the School of Engineering of University of Udine; see \href{https://opthub.uniud.it/problem/timetabling/gbac}{opthub.uniud.it}.
The structure of a data file in JSON format is as follows:

\begin{json}
{
  "nYears": 3,
  "nPeriodsPerYear": 3,
  "loadBounds": {"min": 2, "max": 6},
  "courseLoads": [6, 5, 5, ..., 1],
  "curricula": ...,
  "precedences": [[121, 3], [121, 6], ...],
  "undesiredPeriods": [[257, 1], [240, 2], ...]
}
\end{json}

\paragraph{Model.}
The \p3 model, in a file `GBACP.py', used for the competition is: 

\begin{boxpy}\begin{python}
@\imp@

nYears, nPeriodsPerYear, loadBounds, courseLoads, curricula, precedences, uPeriods = data
nPeriods, nCourses, nCurricula = nYears * nPeriodsPerYear, len(courseLoads), len(curricula)
loadRange = range(loadBounds.min, loadBounds.max + 1)

max_load = sum(courseLoads)
total_load = [sum(courseLoads[i] for i in c) for c in curricula]
ideal_floor = [total_load[c] // nPeriods for c in range(nCurricula)]
ideal_ceil = [ideal_floor[c] + (0 if total_load[c] % nPeriods == 0 else 1)
   for c in range(nCurricula)]
distinctCurricula = [curricula[i] for i in range(nCurricula)
   if all(curricula[j] != curricula[i] for j in range(i))]

# x[i] is the period for the ith course
x = VarArray(size=nCourses, dom=range(nPeriods))

# y[c][p] is the load in period p of curriculum c
y = VarArray(size=[nCurricula, nPeriods], dom=range(max_load + 1))

# d[c][p] is the delta between the ideal and effective loads in period p of curriculum c
d = VarArray(size=[nCurricula, nPeriods], dom=range(max_load + 1))

satisfy(
   # respecting authorized loads of courses for all periods and curricula
   [Cardinality(x[crm], occurrences={p: loadRange for p in range(nPeriods)})
      for crm in distinctCurricula],

   # respecting prerequisites
   [x[i] < x[j] for i, j in precedences],

   # computing loads
   [BinPacking(x[crm], sizes=courseLoads[crm], loads=y[c]) for c,crm in enumerate(curricula)]
)

if not variant():
   satisfy(
      # computing deltas
      d[c][p] == Maximum(y[c][p] - ideal_ceil[c], ideal_floor[c] - y[c][p])
         for c in range(nCurricula) for p in range(nPeriods)
   )

elif variant("table"):
   T = [[(max(v - ideal_ceil[c], ideal_floor[c] - v), v) for v in range(max_load + 1)]
      for c in range(nCurricula)]

   satisfy(
      # computing deltas
      (d[c][p], y[c][p]) in T[c] for c in range(nCurricula) for p in range(nPeriods)
   )

minimize(
   # minimizing preference violations and unbalanced loads
   Sum(d[c][p] * d[c][p] for c in range(nCurricula) for p in range(nPeriods))
   + Sum(x[i] == v + k*nPeriodsPerYear for i, v in uPeriods for k in range(nPeriodsPerYear))
)
\end{python}\end{boxpy}

This model involves 3 arrays of variables, and 5 types of constraints: \gb{Cardinality}, \gb{Intension}, \gb{BinPacking}, \gb{Maximum} and \gb{Extension}.
Actually, depending on the chosen variant, either \gb{Maximum} constraints are posted, or \gb{Extension} constraints are posted.
Note that \verb!x[crm]! is a shortcut for \verb![x[i] for i in crm]!. 

A series of $2*10$ instances has been selected for the competition (10 per variant).
For generating an \x3 instance (file), you can execute for example:
\begin{command}
python GBACP.py -data=UD01.json
python GBACP.py -data=UD01.json -variant=table
\end{command}
where `UD01.json' is a data file in JSON format.
Note that when you omit to write `-variant=table', you get the main variant.

\subsection{Generalized MKP}

\paragraph{Description {\small (excerpt from Wikipedia)}.}
In this variation of the knapsack problem, the weight of knapsack item $i$ is given by a D-dimensional vector $w_i=(w_{i_1}, \dots, w_{i_D})$ and the knapsack has a D-dimensional capacity vector $(W_{1},\dots ,W_{D})$.
The target is to maximize the sum of the values of the items in the knapsack so that the sum of weights in each dimension $d$ does not exceed $W_{d}$. 
See \href{https://en.wikipedia.org/wiki/Knapsack_problem}{Wikipedia}.

\paragraph{Data.}
As an illustration of data specifying an instance of this problem, we have:

\begin{json}
{
  "profits": [504, 803, 667, ..., 632],
  "wmatrix": [
    [42, 41, 523, ..., 298],
    [509, 883, 229, ..., 850],
    [806, 361, 199, ..., 447],
    [404, 197, 817, ..., 322],
    [475, 36, 287, ..., 635]],
  "capacities": [11927, 13727, 11551, 13056, 13460],
  "pmatrix": [
    [866, 690, 813, ..., 717],
    [1022, 959, 510, ..., 813],
    [866, 1022, 654, ..., 625],
    [855, 853, 811, ...4, 932],
    [826, 654, 636, ..., 640]]
}
\end{json}

Data are from the Chu and Beasley series coming from  \href{https://www.researchgate.net/publication/271198281_Benchmark_instances_for_the_Multidimensional_Knapsack_Problem}{ResearchGate}.

\paragraph{Model.}
The \p3 model, in a file `GeneralizedMKP.py', used for the competition is: 

\begin{boxpy}\begin{python}
@\imp@

profits, WM, capacities, PM = data  # Weight and Profit Matrices  
nItems, nBins = len(profits), len(capacities)

# x[i] is 1 if the item i is packed
x = VarArray(size=nItems, dom={0, 1})

# w[j] si the total weight in the jth bin
w = VarArray(size=nBins, dom=lambda j: range(capacities[j] + 1))

# z is the general profit
z = Var(range(sum(profits) + 1))

satisfy(
   [Knapsack(x, weights=W, wlimit=w[j], profits=PM[j]) >= z for j, W in enumerate(WM)],

   # computing the objective value
   z == profits * x
)

maximize(
   # maximizing the profit of packed items
   z
)
\end{python}\end{boxpy}

This problem involves 2 arrays of variables, a stand-alone variable and 2 types of constraints: \gb{Knapsack}, and \gb{Sum}.
A series of 15 instances has been selected for the competition.
For generating an \x3 instance (file), you can execute for example:
\begin{command}
python GeneralizedMKP.py -data=OR05x100-25-1.json
\end{command}
where `OR05x100-25-1.json' is a data file in JSON format.

\subsection{HC Pizza}

This is Pizza Practice Problem for \href{https://www.academia.edu/31537057/Pizza_Practice_Problem_for_Hash_Code_2017}{Hash Code 2017}.

\paragraph{Description.}

From HC'17: ``The pizza corresponds to a 2-dimensional grid of $n$ rows and $m$ columns.
Each cell of the pizza contains either mushroom or tomato.
A slice of pizza is a rectangular section of the pizza delimited by two rows and two columns, without holes.
The slices we want to cut out must contain at least $L$ cells of each ingredient and at most $H$ cells of any kind in total.
The slices being cut out cannot overlap, and do not need to cover the entire pizza.
The goal is to cut correct slices out of the pizza maximizing the total number of cells in all slices.''

\paragraph{Data.}
As an illustration of data specifying an instance of this problem, we have:

\begin{json}
{
  "minIngredients": 2,
  "maxSize": 6,
  "pizza": [
    [1, 1, 0, 1, 1, 1, 1, 1, 1, 0],
    [0, 1, 0, 0, 1, 0, 1, 0, 0, 1],
    ..., 
    [1, 0, 0, 1, 0, 1, 1, 0, 0, 0]
  ]
}
\end{json}

\paragraph{Model.}
The \p3 model, in a file `HCPizza.py', used for the competition is: 

\begin{boxpy}\begin{python}
@\imp@

minIngredients, maxSize, pizza = data
n, m = len(pizza), len(pizza[0])  # nRows and nColumns
patterns = [(i, j) for i in range(1, min(maxSize, n)+1) for j in range(1, min(maxSize, m)+1) if 2 * minIngredients <= i * j <= maxSize]
nPatterns = len(patterns)

def possible_slices():
   _overlaps = [[[] for _ in range(m)] for _ in range(n)]
   _possibles = [[[False for _ in range(nPatterns)] for _ in range(m)] for _ in range(n)]
  for i, j, k in product(range(n), range(m), range(nPatterns)):
     height, width = patterns[k][0], patterns[k][1]
     n_mushrooms, n_tomatoes = 0, 0
     for ib, jb in product(range(i, min(i + height, n)), range(j, min(j + width, m))):
        if pizza[ib][jb] == 0:
           n_mushrooms += 1
        else:
           n_tomatoes += 1
     if n_mushrooms >= minIngredients and n_tomatoes >= minIngredients:
        _possibles[i][j][k] = True
        for ib, jb in product(range(i, min(i + height, n)), range(j, min(j + width, m))):
           _overlaps[ib][jb].append((i, j, k))
  return _overlaps, _possibles

overlaps, slices = possible_slices()

def pattern_size(i, j, k):
    return (min(i + patterns[k][0], n) - i) * (min(j + patterns[k][1], m) - j)


# x[i][j][k] is 1 iff the slice with left top cell at (i,j) and pattern k is selected
x = VarArray(size=[n, m, nPatterns],
              dom=lambda i, j, k: {0, 1} if slices[i][j][k] else None)

# s[i][j][k] is the size of the slice with left top cell at (i,j) and pattern k (0 if the slice is not selected)
s = VarArray(size=[n, m, nPatterns],
              dom=lambda i, j, k: {0, pattern_size(i, j, k)} if slices[i][j][k] else None)

# z is the number of selected pizza cells
z = Var(dom=range(n * m + 1))

satisfy(
   # computing sizes of selected slices
   [(x[i][j][k], s[i][j][k]) in {(0, 0), (1, pattern_size(i, j, k))}
      for i, j, k in product(range(n), range(m), range(nPatterns)) if slices[i][j][k]],

   # ensuring that no two slices overlap
   [Sum(x[t]) <= 1 for i in range(n) for j in range(m) if len(t:=overlaps[i][j]) > 1],

   # computing the number of selected pizza cells
   Sum(s) == z
)

maximize(
   # maximizing the number of selected pizza cells
   z
)   
\end{python}\end{boxpy}

This problem involves 2 arrays of variables, 1 stand-alone variable and 2 types of constraints \gb{Extension} and \gb{Sum}.
A series of 10 (randomly generated) instances has been selected for the competition.
For generating an \x3 instance (file), you can execute for example:

\begin{command}
python HCPizza.py -data=10-10-2-6.json
\end{command}

where `10-10-2-6.json' is a data file in JSON format.

Or, you can use the random generator `HCPizza\_Random.py':
\begin{command}
python HCPizza.py -parser=HCPizza_Random.py 20 20 2 8 2
\end{command}
Note that for saving data in JSON files, you can add the option `-export' (or `-dataexport').

\subsection{Hoist Scheduling Problem (HSP)}

\paragraph{Description.}

From \cite{AE_efficient}: ``we consider a facility with a single handling resource
(a hoist). The hoist has to perform a sequence of moves in order to
accomplish a set of jobs, with varying processing requirements, while
satisfying processing and transport resource constraints. The objective
is to determine a feasible schedule (i.e., a sequence) that minimizes
the total processing time of a set of jobs (i.e., the makespan), while, at the same time, satisfying surface treatment constraints.''
See also \cite{WY_new}.

\paragraph{Data.}
As an illustration of data specifying an instance of this problem, we have:

\begin{json}
{
  "cuves": [[50, 114], [80, 184], [60, 145], [80, 191], [28, 67]], 
  "f": 12,
  "e": 8
}
\end{json}

\paragraph{Model.}
The \p3 model, in a file `Hsp.py', used for the competition is: 

\begin{boxpy}\begin{python}
@\imp@

intervals, ld, ud = data  # loaded and unloaded time for one move 

min_dips, max_dips = zip(*intervals)  # min and max dip durations
nTanks = len(intervals) + 1
horizon = sum(min_dips) + ld * nTanks + ud * nTanks + 1

# t[i] is the time when the product is picked from the ith tank
t = VarArray(size=nTanks, dom=range(horizon - ld - ud))

# o[i] is the order in which the picking operation of the ith tank is executed
o = VarArray(size=nTanks, dom=range(nTanks))

# d[i] is the dip duration in the ith tank (plus ld)
d = VarArray(size=nTanks, dom=lambda i: range(min_dips[i - 1] + ld, max_dips[i - 1] + 1 + ld) if i != 0 else {0})  # i-1 because of special tank 0

# time of the cycle
z = Var(range(nTanks * (ld + ud), horizon))

def duration_limit(i, j, delta):
   return ld + ud * (abs(j - i) + delta)

satisfy(
   # we start the cycle by putting a product in the first tank
   [t[0] == 0, o[0] == 0],

   # taking into account the time of going back to tank 0
   [t[i] <= horizon - ld - (i + 1) * ud for i in range(1, nTanks)],

   # operations are executed in some order
   AllDifferent(o),

   # order of picking operations must be compatible with their picking times
   [(o[i] < o[j]) == (t[i] < t[j]) for i, j in combinations(nTanks, 2)],

   # ensuring a minimal duration between any two tanks
   NoOverlap(origins=t, lengths=ld + min(ud, min(min_dips))),

   # computing the time of the cycle
   z == Maximum(t[i] + ld + (i + 1) * ud for i in range(1, nTanks)),

   # ensuring a duration limit between any two tanks
   [
      (
         If(o[j] - o[i] == 1, Then=t[j] - t[i] >= duration_limit(i, j, -1)),
         If(o[i] - o[j] == 1, Then=t[i] - t[j] >= duration_limit(i, j, +1))
      ) for i, j in combinations(nTanks, 2)
   ]    
)

if not variant():
   satisfy(
      # computing the dip duration of each product
      d[i] == ift(o[i] < o[i-1], z + t[i] - t[i-1], t[i] - t[i-1]) for i in range(1, nTanks)
   )
else:
   assert variant() in ("aux", "table")

   # g[i] is the gap (distance) between t[i] and t[i+1]
   g = VarArray(size=nTanks - 1, dom=range(-horizon, horizon))

   satisfy(
      # reducing the domain of auxiliary variables
      (
         g[i] < horizon - ld - (i + 2) * ud,
         g[i] > -horizon + ld + (i + 2) * ud
      ) for i in range(1, nTanks - 1)
   )
            
   if variant("aux"):

      satisfy(
         # computing the gap/distance between the picking time concerning two successive tanks
         [g[i] == t[i + 1] - t[i] for i in range(nTanks - 1)],

         # computing the dip duration of each product
         [d[i] == ift(g[i - 1] < 0, z + g[i - 1], g[i - 1]) for i in range(1, nTanks)]
      )

   elif variant("table"):

      T1 = [(v2 - v1, v2, v1) for v2 in t[2].dom for v1 in t[1].dom if v2 - v1 in g[1].dom]
      T2 = [[(vt + vg if vg < 0 else vg, vg, vt) for vg in g[i - 1].dom for vt in z.dom
              if (vt + vg if vg < 0 else vg) in d[i].dom] for i in range(1, nTanks)]

      satisfy(
         # computing the gap/distance between two successive tanks
         [(g[i], t[i + 1], t[i]) in T1 for i in range(nTanks - 1)],

         # computing the dip duration of each product
         [(d[i], g[i - 1], z) in T2[i - 1] for i in range(1, nTanks)]
      )

minimize(
   #  minimizing the time of the cycle
   z
)
\end{python}\end{boxpy}

This problem involves 3 arrays of variables, a stand-alone variable ($z$) and 5 types of constraints: \gb{Intension}, \gb{AllDifferent}, \gb{NoOverlap}, \gb{Maximum} and \gb{Extension}.
Actually, the constraints \gb{Extension} are only posted when the chosen variant is `table'.
A series of $6*3$ instances has been selected for the competition (6 per variant).
For generating an \x3 instance (file), you can execute for example:
\begin{command}
python Hsp.py -data=10405.json
python Hsp.py -data=10405.json -variant=aux
python Hsp.py -data=10405.json -variant=table
\end{command}
where `10405.json' is a data file in JSON format.

Note that when you omit to write `-variant=aux' or `-variant=table', you get the main variant.

\subsection{Kidney Exchange}

\paragraph{Description.}
From \href{https://en.wikipedia.org/wiki/Optimal_kidney_exchange}{Wikipedia}: ``Optimal kidney exchange is an optimization problem faced by programs for kidney paired donations (also called Kidney Exchange Programs).
Such programs have large databases of patient-donor pairs, where the donor is willing to donate a kidney in order to help the patient, but cannot do so due to medical incompatibility.
The centers try to arrange exchanges between such pairs.
For example, the donor in pair A donates to the patient in pair B, the donor in pair B donates to the patient in pair C, and the donor in pair C donates to the patient in pair A.
The objective is to find an optimal arrangement of such exchanges.''

\paragraph{Data.}
As an illustration of data specifying an instance of this problem, we have:

\begin{json}
{
  "weights": [
    [-1, -1, -1, -1, 1, -1, 1, -1, -1, -1, -1, -1, -1, -1, -1, -1, 0, -1],
    [-1, -1, -1, -1, -1, -1, -1, -1, -1, -1, -1, -1, 1, -1, 1, -1, 0, -1],
    ...,
    [-1, -1, -1, -1, -1, -1, -1, -1, -1, -1, -1, -1, -1, -1, -1, -1, -1, -1]],
  "k": 4
}
\end{json}

The data files used for the competition were generated by John Dickerson (see \cite{DPS_optimizing}), and are available at \href{https://www.preﬂib.org/dataset/00036}{PrefLib}.

\paragraph{Model.}
The \p3 model, in a file `KidneyExchange.py', used for the competition is: 

\begin{boxpy}\begin{python}
@\imp@

weights, k = data
n = len(weights)

# x[i] is the successor node of node i (in the cycle where i belongs)
x = VarArray(size=n, dom=range(n))

# y[i] is the cycle (index) where the node i belongs
y = VarArray(size=n, dom=range(n))

satisfy(
   AllDifferent(x),

   # ensuring correct cycles
   [y[i] == y[x[i]] for i in range(n)],

   # disabling infeasible arcs
   [x[i] != j for i in range(n) for j in range(n) if i != j and weights[i][j] < 0],

   # each cycle has k as maximum length
   BinPacking(y, sizes=1) <= k,

   # tag(symmetry-breaking)
   Precedence(y)
)

maximize(
   # maximizing the sum of arc weights of selected cycles
   Sum(weights[i][x[i]] for i in range(n))
)
\end{python}\end{boxpy}

Here, the model is equivalent to the one proposed by Edward Lam and Vicky H. Mak-Hau at the Minizinc challenge 2019.
See also \cite{M_kep}.
This model involves 2 arrays of variables and 5 types of constraints: \gb{AllDifferent}, \gb{Element}, \gb{Intension} \gb{BinPacking} and \gb{Precedence}.
A series of 18 instances has been selected for the competition (coming from data at \href{https://www.preﬂib.org/dataset/00036}{PrefLib}).
For generating an \x3 instance (file), you can execute for example:
\begin{command}
python KidneyExchange.py -data=4-001.json
\end{command}
where `4-001.json' is a data file in JSOn format.

\subsection{K-Median Problem}

\paragraph{Description.}

From \href{https://en.wikipedia.org/wiki/K-medians_clustering}{Wikipedia}: ``The k-median problem (with respect to the 1-norm) is the problem of finding $k$ centers such that the clusters formed by them are the most compact.
Formally, given a set of data points, the $k$ centers $c_i$ are to be chosen so as to minimize the sum of the distances from each data point to the nearest $c_i$.''
See also \cite{B_note}, as well as the results in Chapter 11 of \cite{HM_constraint}. 

\paragraph{Data.}
As an illustration of data specifying an instance of this problem, we have:

\begin{json}
{
  "distances": [
    [0, 30, 76, ..., 88],
    [30, 0, 46, ..., 118],
    ...
    [88, 118, 98, ..., 0]
  ],
  "k": 5
}
\end{json}

\paragraph{Model.}
The \p3 model, in file `KMedian.py', used for the competition is: 

\begin{boxpy}\begin{python}
@\imp@

distances, k = data
n = len(distances)

# x[i] is the ith selected node
x = VarArray(size=k, dom=range(n))

satisfy(
   # selected nodes must be all different
   AllDifferent(x),

   # tag(symmetry-breaking)
   Increasing(x, strict=True)
)

if not variant():
   minimize(
      # minimizing the minimal distances between nodes and the selected ones
      Sum(Minimum(distances[c][j] for c in x) for j in range(n))
   )

elif variant("aux"):

   # d[i][j] is the distance between the ith selected node and the jth node
   d = VarArray(size=[k, n], dom=distances)

   satisfy(
      # computing distances
      d[i][j] == distances[:, j][x[i]] for i in range(k) for j in range(n)
   )

   minimize(
      # minimizing the minimal distances between nodes and the selected ones
      Sum(Minimum(d[:, j]) for j in range(n))
   )
\end{python}\end{boxpy}

For the variant `aux' (used for the competition), this model involves 2 arrays of variables and 5 types of constraints: \gb{AllDifferent}, \gb{Increasing}, \gb{Element}, \gb{Minimum} and \gb{Sum}.
A series of $15$ instances has been selected for the competition, coming from the data set `p-median - uncapacitated' at \href{http://people.brunel.ac.uk/~mastjjb/jeb/orlib/pmedinfo.html}{OR-Library}.
For generating an \x3 instance (file), you can execute for example:
\begin{command}
python KMedian.py -data=pmed01.json -variant=aux 
\end{command}
where `pmed01.json' is a data file in JSON format.

\subsection{Large Scale Scheduling}

\paragraph{Description.}
Large scale scheduling instances have been built to show the interest of a scalable time-table filtering algorithm for the \gb{Cumulative} Constraint \cite{GHS_simple}.

\paragraph{Data.}
As an illustration of data specifying an instance of this problem, we have:

\begin{json}
{
  "limit": 100,
  "durations": [1460, 6048, 8129, ..., 12471],
  "heights": [17, 6, 40, ..., 19]
}
\end{json}

Data files can be found at \href{http://becool.info.ucl.ac.be/pub/resources/large-scale-scheduling-instances.zip}{BeCool - Belgian Constraints Group @ UCLouvain}.

\paragraph{Model.}
The \p3 model, in a file `LargeScaleScehduling.py', used for the competition is: 

\begin{boxpy}\begin{python}
@\imp@

limit, durations, heights = data
nTasks = len(durations)

horizon = sum(durations) + 1  # Trivial upper bound on the horizon

# x[i] is the starting time of the ith task
x = VarArray(size=nTasks, dom=range(horizon))

satisfy(
   # resource cumulative constraint
   Cumulative(origins=x, lengths=durations, heights=heights) <= limit
)

minimize(
   Maximum(x[i] + durations[i] for i in range(nTasks))
)
\end{python}\end{boxpy}

This model involves 1 array of variables and 2 types of constraints: \gb{Cumulative}, and \gb{Maximum} (in the objective).
A series of 9 instances has been selected for the competition.
For generating an \x3 instance (file), you can execute for example:
\begin{command}
python LargeScaleScheduling.py -data=03200-0.json
\end{command}
where `03200-0.json' is a data file in JSON format.

\subsection{Progressive Party}

This is Problem \href{https://www.csplib.org/Problems/prob013}{013} on CSPLib.

\paragraph{Description {\small (excerpt from CSPLib)}.}
``The problem is to timetable a party at a yacht club. Certain boats are to be designated hosts, and the crews of the remaining boats in turn visit the host boats for several successive half-hour periods.
The crew of a host boat remains on board to act as hosts while the crew of a guest boat together visits several hosts. Every boat can only hold a limited number of people at a time (its capacity) and crew sizes are different.
The total number of people aboard a boat, including the host crew and guest crews, must not exceed the capacity. A table with boat capacities and crew sizes can be found below; there were six time periods.
A guest boat cannot not revisit a host and guest crews cannot meet more than once. The problem facing the rally organizer is that of minimizing the number of host boats.''

\paragraph{Data.}
As an illustration of data specifying an instance of this problem, we have:

\begin{json}
{
  "nPeriods": 5,
  "boats": [
    {"capacity": 6, "crewSize": 2},
    {"capacity": 8, "crewSize": 2},
    ...,
    {"capacity": 10, "crewSize": 3}
  ]
}
\end{json}

\paragraph{Model.}
The \p3 model, partly developed from \cite{SBHW_progressive}, in a file `ProgressiveParty.py', used for the competition is: 

\begin{boxpy}\begin{python}
@\imp@

nPeriods, boats = data
nBoats = len(boats)
capacities, crews = zip(*boats)

def minimal_number_of_hosts():
   nPersons = sum(crews)
   cnt, acc = 0, 0
   for capacity in sorted(capacities, reverse=True):
      if acc >= nPersons:
         return cnt
      acc += capacity
      cnt += 1

# h[b] indicates if the boat b is a host boat
h = VarArray(size=nBoats, dom={0, 1})

# s[b][p] is the scheduled (visited) boat by the crew of boat b at period p
s = VarArray(size=[nBoats, nPeriods], dom=range(nBoats))

# g[b1][p][b2] is 1 if s[b1][p] = b2
g = VarArray(size=[nBoats, nPeriods, nBoats], dom={0, 1})

satisfy(
   # identifying host boats
   [h[b] == (s[b][p] == b) for b in range(nBoats) for p in range(nPeriods)],

   # identifying host boats (from visitors)
   [h[s[b][p]] == 1 for b in range(nBoats) for p in range(nPeriods)],

   # channeling variables from arrays s and g
   [Channel(g[b][p], s[b][p]) for b in range(nBoats) for p in range(nPeriods)],

   # boat capacities must be respected
   [g[:, p, b] * crews <= capacities[b] for b in range(nBoats) for p in range(nPeriods)],

   # a guest boat cannot revisit a host
   [AllDifferent(s[b], excepting=b) for b in range(nBoats)],
  
   # guest crews cannot meet more than once
   [Sum(s[b1][p] == s[b2][p] for p in range(nPeriods)) <= 1
     for b1, b2 in combinations(nBoats, 2)],

   # tag(redundant-constraint)
   # ensuring a minimum number of hosts
   Sum(h) >= minimal_number_of_hosts()
)

minimize(
   # minimizing the number of host boats
   Sum(h)
)
\end{python}\end{boxpy}

This problem involves 3 arrays of variables and 5 types of constraints: \gb{Intension}, \gb{Element}, \gb{Channel}, \gb{Sum} and \gb{AllDifferent}.
A series of 7 instances has been selected for the competition.
For generating an \x3 instance (file), you can execute for example:
\begin{command}
python ProgressiveParty.py -parser=ProgressiveParty_rally-red.py 42 12
\end{command}
where `ProgressiveParty\_rally-red.py' is a generator based on the instance `rally-red' that allows us to specify the number of boats and the number of periods.
Note that for saving data in JSON files, you can add the option `-export' (or `-dataexport').

\subsection{Pigment Sequencing Problem (PSP)}

This is a particular case of the Discrete Lot Sizing Problem (DLSP); see Problem \href{https://www.csplib.org/Problems/prob058}{058} on CSPLib.

\paragraph{Description {\small (excerpt from CSPLib)}.}

``Discrete Lot Sizing and Scheduling Problem (DLSP) is a production planning problem which consists of determining a minimal cost production schedule (production costs, setup costs, changeover costs, stocking costs, etc.), such that machine capacity restrictions are not violated, and demand for all products is satisfied. The planning horizon is discrete and finite.
The variant PSP is a multi-item, single machine problem with capacity of production limited to one per period.
There are storage costs and sequence-dependent changeover costs, respecting the triangle inequality.
Each order consisting of one unit of a particular item has a due date and must be produced at latest by its due date.
The stocking (inventory) cost of an order is proportional to the number of periods between the due date and the production period. The changeover cost $q_{i,j}$ is induced when passing from the production of item $i$ to another one $j$ with $q_{i,i}=0, \forall i$.
The objective is to assign a production period for each order respecting its due date and the machine capacity constraint so as to minimize the sum of stocking costs and changeover costs.''

See \cite{HSW_item,GS_lns}.

\paragraph{Data.}
As an illustration of data specifying an instance of this problem, we have:

\begin{json}
{
  "nOrders": 50,
  "changeCosts": [
    [0, 45, 10, 34, 36, 37, 10, 13, 35, 26],
    [18, 0, 47, 32, 30, 32, 49, 30, 40, 31],
    ...,
    [26, 21, 34, 22, 48, 45, 31, 25, 22, 0]
  ],
  "stockingCosts": [21, 38, 25, 10, 48, 42, 22, 24, 44, 38],
  "demands": [
    [0, 0, 0, ..., 1],
    [0, 0, 0, ..., 1],
    ...,
    [0, 0, 0, ..., 0]
  ]
}
\end{json}

\paragraph{Model.}

A first \p3 model, in a file `PSP1.py', used for the competition is: 

\begin{boxpy}\begin{python}
@\imp@
    
nOrders, changeCosts, stockingCosts, demands = data
nItems, horizon = len(demands), len(demands[0])

# x[t][i] is 1 when item i is produced at time t
x = VarArray(size=[horizon, nItems], dom={0, 1})

# y[t][i] is 1 if the machine is ready to produce i at time t
y = VarArray(size=[horizon, nItems], dom={0, 1})

# c[t][i] is 1 if the configuration changes from i to j at time t
c = VarArray(size=[horizon, nItems, nItems], dom=lambda t, i, j: {0, 1} if t != 0 and i != j else None)

# s[t][i] is the number of items of type i stored at time t
s = VarArray(size=[horizon + 1, nItems], dom=lambda t, i: range(sum(demands[i]) + 1))

satisfy(
   # the stock of every item is empty at startup
   s[0] == 0,

   # when an item is produced, it is either delivered or stocked for later delivery
   [x[t][i] + s[t][i] == demands[i][t] + s[t + 1][i] for t in range(horizon)
     for i in range(nItems)],

   # consistency between production and machine setup
   [x[t][i] <= y[t][i] for t in range(horizon) for i in range(nItems)],

   # only 1 unit of one item is produced at each time
   [Sum(y[t]) == 1 for t in range(horizon)],

   # consistency between machine setup and changeover
   [c[t][i][j] >= y[t - 1][i] + y[t][j] - 1 for t in range(1, horizon) for i in range(nItems) for j in range(nItems) if i != j]
)

minimize(
   Sum(changeCosts[i][j] * Sum(c[:, i, j])
     for i in range(nItems) for j in range(nItems) if i != j)
   + Sum(stockingCosts[i] * Sum(s[:, i]) for i in range(nItems))
)
\end{python}\end{boxpy}

A second \p3 model, in a file `PSP2.py', used for the competition is: 

\begin{boxpy}\begin{python}
@\imp@
    
nOrders, changeCosts, stockingCosts, demands = data
nItems, horizon = len(demands), len(demands[0])

required = [[sum(demands[i][:t + 1]) for t in range(horizon)] for i in range(nItems)]

# x[t] is the item produced at time t
x = VarArray(size=horizon, dom=range(nItems))

# p[i][t] is 1 if the ith item is produced at time t
p = VarArray(size=[nItems, horizon], dom={0, 1})

# z[t] is the changeover cost incurred at time t
z = VarArray(size=horizon - 1, dom=changeCosts)

satisfy(
   # channeling variables
   [p[i][t] == (x[t] == i) for i in range(nItems) for t in range(horizon)],

   # ensuring that deadlines of demands are respected
   [Sum(p[i][:t + 1]) >= required[i][t] for i in range(nItems) for t in range(horizon)
     if t == 0 or required[i][t - 1] != required[i][t]],

   # computing changeover costs
   [z[t] == changeCosts[x[t], x[t + 1]] for t in range(horizon - 1)],

   # tag(redundant-constraints)
   [Count(x, value=i) >= required[i][-1] for i in range(nItems)]
)

minimize(
   Sum(stockingCosts[i] * (Sum(p[i][:t + 1]) - required[i][t])
     for i in range(nItems) for t in range(horizon))
   + Sum(z)
)
\end{python}\end{boxpy}

These models can be, e.g., compared to the one written by Xavier Gillard; see \href{https://github.com/xgillard/mznlauncher}{GitHub}.
The second model above involves 3 arrays of variables and 4 types of constraints: \gb{Intension}, \gb{Sum}, \gb{Element} and \gb{Count}.
A series of $8*2$ instances has been selected for the competition (8 per model), from data files available in the \href{https://github.com/xgillard/ijcai_22_DDLNS}{Supplementary Material} of paper \cite{GS_lns}.
For generating an \x3 instance (file), you can execute for example:
\begin{command}
python PSP1.py -data=001.json
python PSP2.py -data=001.json  
\end{command}
where `001.json' is a data file in JSON format.

\subsection{Resource Investment Problem (RIP)}

The Resource Investment Problem (RIP) is also known as the Resource Availability Cost Problem (RACP).

\paragraph{Description}

From \href{https://www.projectmanagement.ugent.be/research/project_scheduling/racp}{OR\&S Research Group at Ghent University}: ``The RIP assumes that the level of renewable resources can be varied at a certain cost and aims at minimizing this total cost of the (unlimited) renewable resources required to complete the project by a pre-specified project deadline.''
See also \cite{KSSZ_mixed}.

\paragraph{Data.}

As an illustration of data specifying an instance of this problem, we have:
\begin{json}
{
  "horizon": 96,
  "costs": [7, 7, 1, 5],
  "jobs": [
    {"duration": 8, "successors": [3, 8, 13], "requirements": [10, 0, 0, 0]},
    {"duration": 1, "successors": [5, 12, 27], "requirements": [0, 1, 0, 0]},
    {"duration": 10, "successors": [6, 10, 14], "requirements": [0, 9, 0, 0]},
    ...,
    {"duration": 1, "successors": [], "requirements": [0, 0, 0, 1]}
  ]
}
\end{json}

\paragraph{Model.}

The \p3 model, in a file `RIP.py', used for the competition is: 

\begin{boxpy}\begin{python}
@\imp@

horizon, costs, tasks = data
durations, successors, requirements = zip(*tasks)
nResources, nTasks = len(costs), len(tasks)
requirements = [[r[k] for r in requirements] for k in range(nResources)]

lb_usage = [max(row) for row in requirements]
ub_usage = [sum(row) for row in requirements]

# s[i] is the starting time of the ith task
s = VarArray(size=nTasks, dom=range(horizon + 1))

# u[k] is the maximal usage (at any time) of the kth resource
u = VarArray(size=nResources, dom=lambda k: range(lb_usage[k], ub_usage[k] + 1))

satisfy(
   # ending tasks before the given horizon
   [s[i] + durations[i] <= horizon for i in range(nTasks)],

   # respecting precedence relations
   [s[i] + durations[i] <= s[j] for i in range(nTasks) for j in successors[i]],

   # cumulative resource constraints
   [Cumulative(origins=s, lengths=durations, heights=requirements[k]) <= u[k]
     for k in range(nResources)]
)

minimize(
   # minimizing weighted usage of resources
   costs * u
)
\end{python}\end{boxpy}

The model involves 2 arrays of variables, and 2 types of constraints: \gb{Intension} and \gb{Cumulative}.
A series of 12 instances has been selected for the competition: they correspond to applying some random perturbation on classical RCPSP instances.
For generating an \x3 instance (file), you can execute for example:
\begin{command}
python RIP.py -data=25-3-j060-01-01.json 
python RIP.py -data=[25,3,j060-30-01] -parser=RIP_Parser.py 
\end{command}
where `25-0-j060-01-01.json' is a data file in JSON format, whereas `j060-30-01' is a data file in tabular text format and `RIP\_Parser.py' is a parser (i.e., a Python file allowing us to load data that are not directly given in JSON format).
Note that for saving data in JSON files, you can add the option `-export' (or `-dataexport').

\subsection{Rule Mining}

This series of instances has been generated by Nicolas Szczepanski from the classification rule mining problem, which is a problem over (discrete and) imbalanced data whose distribution greatly varies over the classes.
A series of 9 instances has been selected for the competition.

\subsection{Sonet}

This is Problem \href{https://www.csplib.org/Problems/prob064}{064} on CSPLib, called Synchronous Optical Networking Problem (SONET).

\paragraph{Description.}

From CSPLib: ``In the SONET problem we are given a set of nodes, and for each pair of nodes we are given the demand (which is the number of channels required to carry network traffic between the two nodes).
The demand may be zero, in which case the two nodes do not need to be connected.
A SONET ring connects a set of nodes.
A node is installed on a ring using a piece of equipment called an add-drop multiplexer (ADM).
Each node may be installed on more than one ring.
Network traffic can be transmitted from one node to another only if they are both installed on the same ring.
Each ring has an upper limit on the number of nodes, and a limit on the number of channels.
The demand of a pair of nodes may be split between multiple rings.
The objective is to minimize the total number of ADMs used while satisfying all demands.'' 

\paragraph{Data.}

As an illustration of data specifying an instance of this problem, we have:
\begin{json}
{
  "n":6,
  "m":10,
  "r":3,
  "connections":[[0,1],[0,2],[0,3],[2,3],[2,5],[4,5]]
}
\end{json}

\paragraph{Model.}
The \p3 model, in a file `Sonet.py', used for the competition is: 

\begin{boxpy}\begin{python}
@\imp@

n, m, r, connections = data

# x[i][j] is 1 if the ith ring contains the jth node
x = VarArray(size=[m, n], dom={0, 1})

T = {tuple(1 if j // 2 == i else ANY for j in range(2 * m)) for i in range(m)}


satisfy(
   [(x[i][conn] for i in range(m)) in T for conn in connections],

   # respecting the capacity of rings
   [Sum(x[i]) <= r for i in range(m)],

   # tag(symmetry-breaking)
   LexIncreasing(x)
)

minimize(
   # minimizing the number of nodes installed on rings
   Sum(x)
)
\end{python}\end{boxpy}

This problem involves 1 array of variables and 3 types of constraints: \gb{Extension}, \gb{Sum} and \gb{LexIncreasing}.
A series of 16 instances has been selected for the competition, from data files corresponding to parameters for Conjure \cite{AFGJMN_conjure}.
For generating an \x3 instance (file), you can execute for example:
\begin{command}
python Sonet.py -data=s3ring03.json
\end{command}
where `s3ring03.json' is a data file in JSON format.

\subsection{Single-Row Facility Layout Problem (SRFLP)}

\paragraph{Description.}

from \cite{CGS_solving}: ``The Single-Row Facility Layout Problem (SRFLP) is an ordering problem considering a set of departments in a facility, with given lengths and pairwise traffic intensities.
Its goal is to find a linear ordering of the departments minimizing the weighted sum of the distances between department pairs''

\paragraph{Data.}

As an illustration of data specifying an instance of this problem, we have:
\begin{json}
{
  "lengths": [40, 20, 50, 30, 10, 50, 80],
  "traffics": [
    [0, 5, 2, 4, 1, 0, 0],
    [5, 0, 3, 0, 2, 2, 2],
    ..., 
    [0, 2, 5, 2, 0, 5, 0]
  ]
}
\end{json}

\paragraph{Model.}
The \p3 model, in a file `SRFLP.py', used for the competition is: 

\begin{boxpy}\begin{python}
@\imp@

lengths, traffics = data
n = len(lengths)

ordered_lengths = sorted(lengths)

pairs = [(i, j) for i, j in combinations(n, 2) if i + 1 < j]

# x[i] is the department at the ith position
x = VarArray(size=n, dom=range(n))

# y[i] is the length of the department at the ith position
y = VarArray(size=n, dom=lengths)

# d[i][j] is the end-to-start distance between the ith and the jth departments
d = VarArray(size=[n, n], dom=lambda i, j: range(sum(ordered_lengths[:j - i - 1]), sum(ordered_lengths[i - j + 1:]) + 1) if i + 1 < j else None)

satisfy(
   # ensuring a linear ordering of the departments
   AllDifferent(x),

   # computing lengths
   [(x[i], y[i]) in [(j, lengths[j]) for j in range(n)] for i in range(n)],

   # computing distances
   [d[i][j] == Sum(y[i + 1:j]) for i, j in pairs]
)

minimize(
   # minimizing weighted end-to-start distances
   Sum(d[i][j] * traffics[x[i], x[j]] for i, j in pairs)
)
\end{python}\end{boxpy}

This model involves 3 arrays of variables and 3 type of constraints: \gb{AllDifferent}, \gb{Extension} and \gb{Sum}.
A series of 15 instances has been selected for the competition, from data files in the \href{https://github.com/vcoppe/csrflp-dd}{Supplemental Material} of Paper \cite{CGS_solving}.
For generating an \x3 instance (file), you can execute for example:
\begin{command}
python SRFLP.py -data=Cl12.json
\end{command}
where `Cl12.json' is a data file in JSON format.

\subsection{TSPTW}

\paragraph{Description.}
Traveling Salesman Problem with Time Windows (TSPTW) is a popular variant of the TSP where the salesman’s customers must be visited within given time windows.

\paragraph{Data.}

As an illustration of data specifying an instance of this problem, we have:
\begin{json}
{
  "distances": [
    [0, 19, 17, ..., 12],
    [19, 0, 10, ..., 31],
    ...,
    [12, 31, 29, ..., 0]
  ],
  "windows": [[0, 408], [62, 68], [181, 205], ..., [275, 300]]
}

\end{json}

\paragraph{Model.}
A first \p3 model, in a file `TSPTW1.py', used for the competition is: 

\begin{boxpy}\begin{python}
@\imp@

distances, windows = data
Earliest, Latest = cp_array(zip(*windows))
horizon = max(Latest) + 1
n = len(distances)

# x[i] is the customer (node) visited in the ith position
x = VarArray(size=n + 1, dom=range(n))

# a[i] is the time when is visited the customer in the ith position
a = VarArray(size=n, dom=range(horizon))

satisfy(
   #  making it a tour while starting and ending at city 0
   [x[0] == 0, x[-1] == 0, a[0] == 0],

   AllDifferent(x[:-1]),

   # enforcing time windows
   [
      [Earliest[x[i]] <= a[x[i]] for i in range(n)],
      [a[x[i]] <= Latest[x[i]] for i in range(n)],
      [a[x[i + 1]] >= a[x[i]] + distances[x[i], x[i + 1]] for i in range(n - 1)]
   ]
)

minimize(
   # minimizing travelled distance
   Sum(distances[x[i], x[(i + 1) % n]] for i in range(n))
)
\end{python}\end{boxpy}

\medskip
A second \p3 model, in a file `TSPTW2.py', used for the competition is: 

\begin{boxpy}\begin{python}
@\imp@

distances, windows = data
horizon = max(latest for (_, latest) in windows) + 1
n = len(distances)

# x[i] is the node succeeding to the ith node
x = VarArray(size=n, dom=range(n))

# a[i] is the time when is visited the ith node
a = VarArray(size=n, dom=lambda i: range(windows[i][0], windows[i][1] + 1))

satisfy(
   #  making it a tour while starting and ending at city 0
   a[0] == 0,

   # avoiding self-loops
   [x[i] != i for i in range(n)],

   # forming a circuit
   Circuit(x),

   # enforcing time windows
   [
      If(
         x[i] != 0,
         Then=a[x[i]] >= a[i] + distances[i][x[i]]
      ) for i in range(n)
   ]
)

minimize(
   # minimizing travelled distance
   Sum(distances[i, x[i]] for i in range(n))
)
\end{python}\end{boxpy}

A series of $2*8$ instances has been selected for the competition (8 per model), from data files available in the \href{https://github.com/xgillard/ijcai_22_DDLNS}{Supplementary Material} of paper \cite{GS_lns}.
For generating an \x3 instance (file), you can execute for example:
\begin{command}
python TSPTW1.py -data=n020w020-1.json
python TSPTW2.py -data=n020w020-1.json
\end{command}
where `n020w020-1.json' is a data file in JSON format.

\chapter{Solvers}

In this chapter, we introduce the solvers and teams having participated to the \x3 Competition 2023.
%When names of solvers are given in italic font, it means that a short description of these solvers are given by authors in the following pages. 

\begin{itemize}
\item ACE (Christophe Lecoutre)
\item BTD, miniBTD (Mohamed Sami Cherif, Djamal Habet, Philippe J\'egou, H\'el\`ene Kanso, Cyril Terrioux)
\item Choco (Charles Prud'homme)
\item CoSoCo (Gilles Audemard)
\item Exchequer (Martin Mariusz Lester)
\item Fun-sCOP (Takehide Soh, Daniel Le Berre, Hidetomo Nabeshima, Mutsunori Banbara, Naoyuki Tamura)
%\item Glasgow (Ciaran McCreesh)
\item MiniCPBP (Gilles Pesant and Auguste Burlats)
\item Mistral (Emmanuel Hebrard and Mohamed Siala)
\item Nacre (Ga\"el Glorian)
\item Picat (Neng-Fa Zhou)
\item RBO, miniRBO (Mohamed Sami Cherif, Djamal Habet, Cyril Terrioux)
\item Sat4j-CSP-PB (extension of Sat4j by Thibault Falque and Romain Wallon)
\item SeaPearl (Max Bourgeat, Axel Navarro, Léo Boisvert, Tom Marty, Louis-Martin Rousseau, Quentin Cappart)
\item toulbar2 (David Allouche et al.) 
\end{itemize}

%\section{BTD}
\addcontentsline{toc}{section}{\numberline{}ACE}
\includepdf[pages=-,pagecommand={\thispagestyle{plain}}]{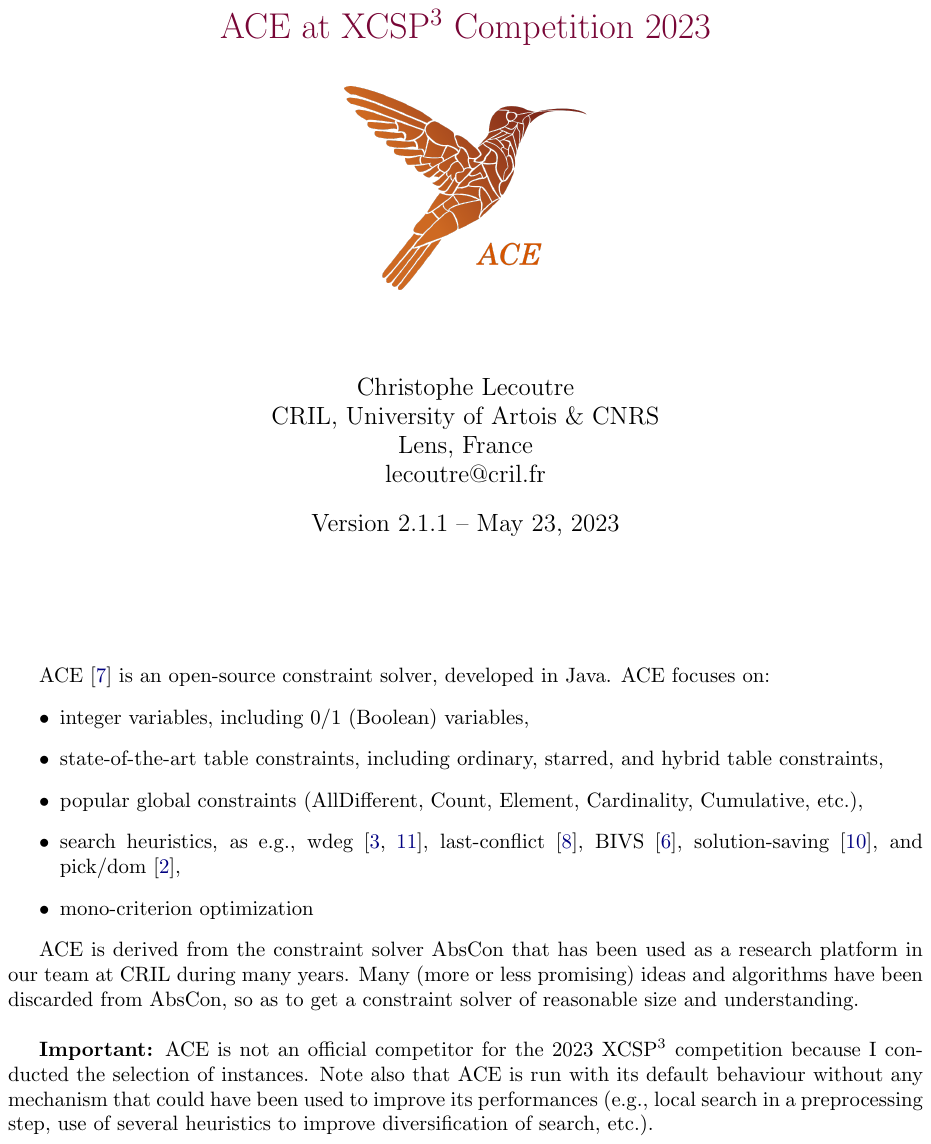}
\addcontentsline{toc}{section}{\numberline{}BTD}
\includepdf[pages=-,pagecommand={\thispagestyle{plain}}]{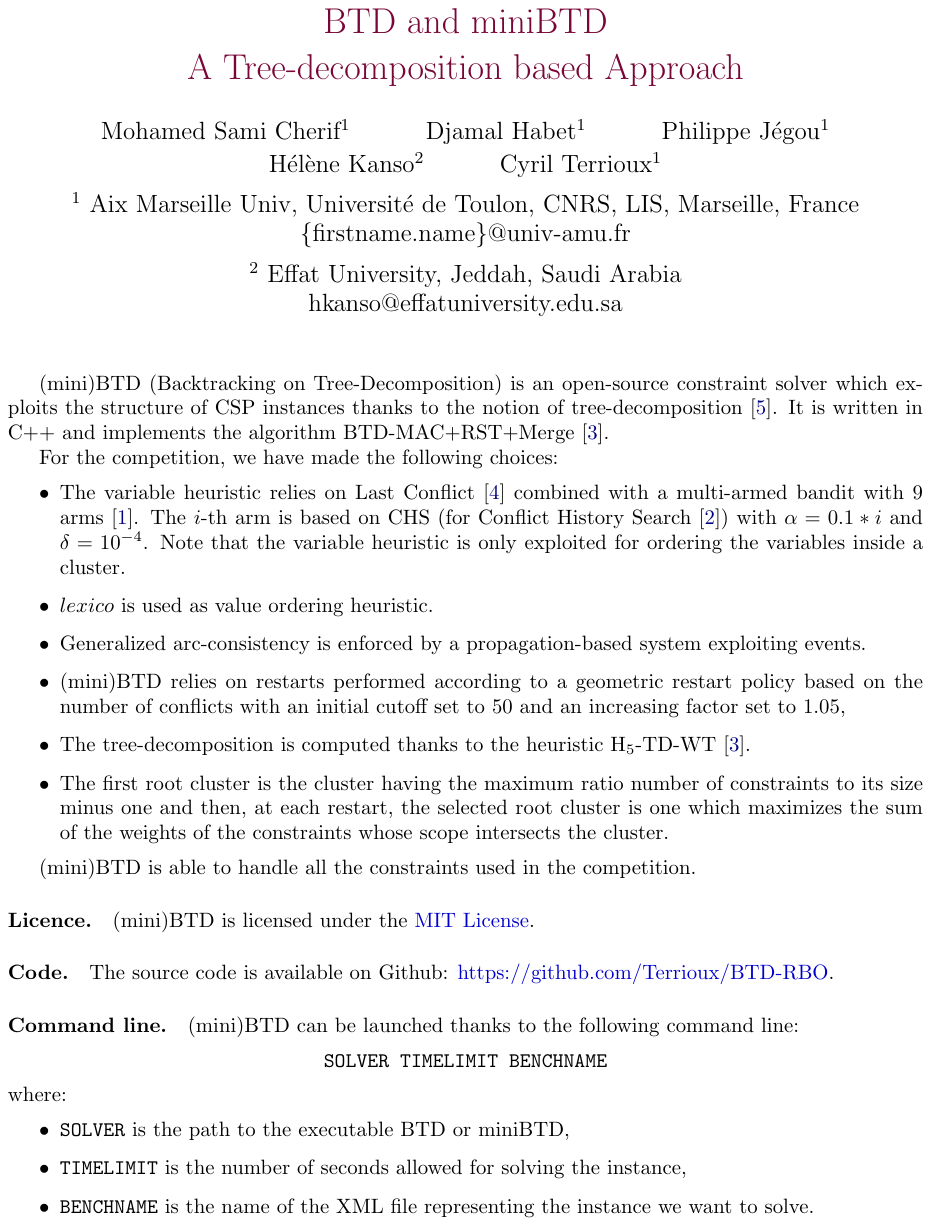}
\addcontentsline{toc}{section}{\numberline{}Choco}
\includepdf[pages=-,pagecommand={\thispagestyle{plain}}]{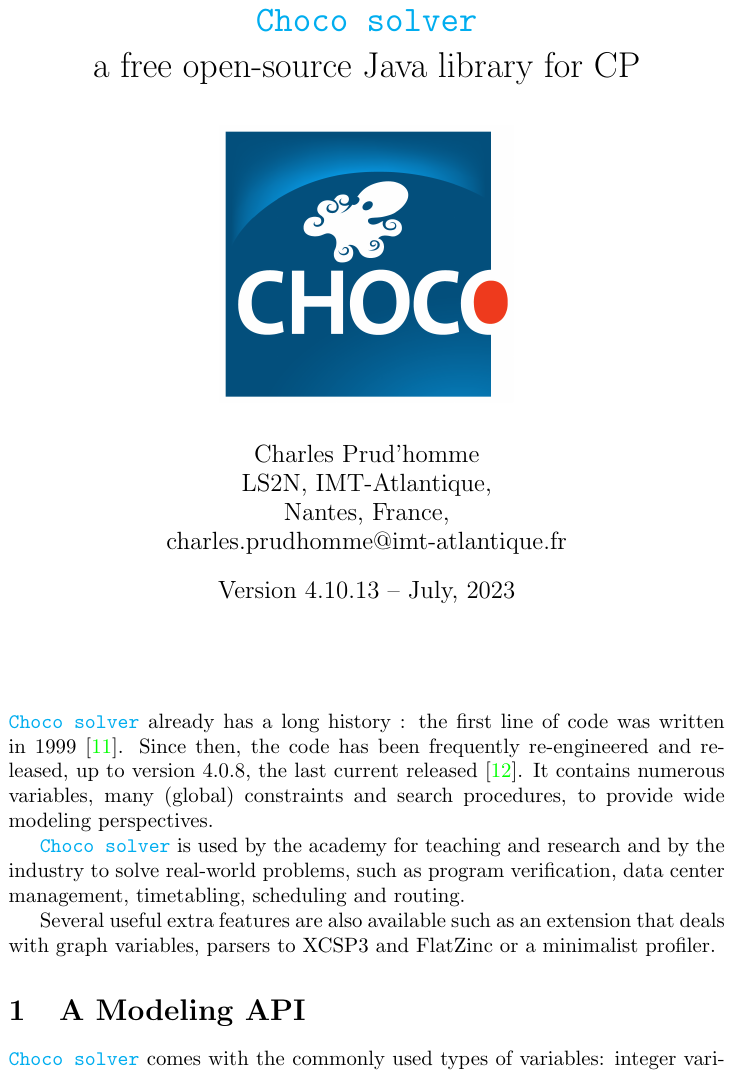}
\addcontentsline{toc}{section}{\numberline{}CoSoCo}
\includepdf[pages=-,pagecommand={\thispagestyle{plain}}]{CoSoCo.pdf}
\addcontentsline{toc}{section}{\numberline{}Exchequer}
\includepdf[pages=-,pagecommand={\thispagestyle{plain}}]{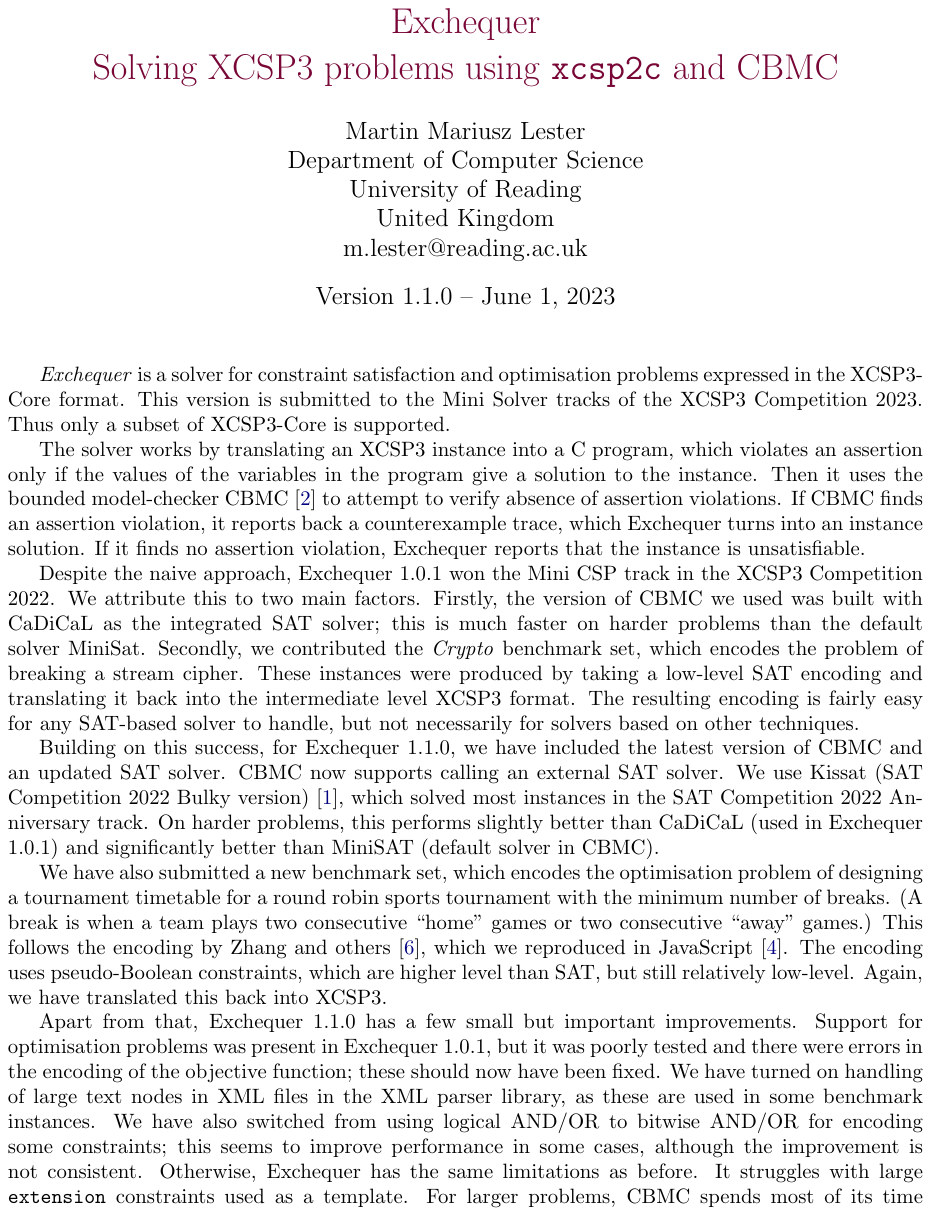}
\addcontentsline{toc}{section}{\numberline{}Fun-sCOP}
\includepdf[pages=-,pagecommand={\thispagestyle{plain}}]{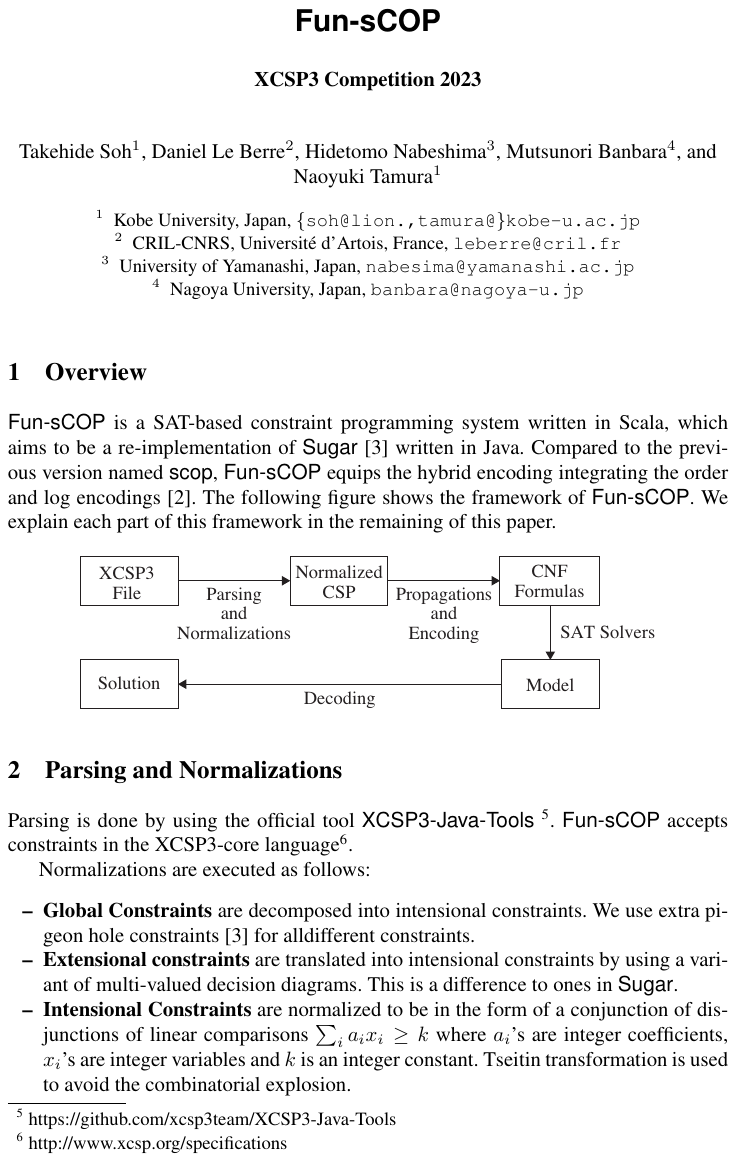}
\addcontentsline{toc}{section}{\numberline{}MiniCPBP}
\includepdf[pages=-,pagecommand={\thispagestyle{plain}}]{miniCPBP.pdf}
\addcontentsline{toc}{section}{\numberline{}Mistral}
\includepdf[pages=-,pagecommand={\thispagestyle{plain}}]{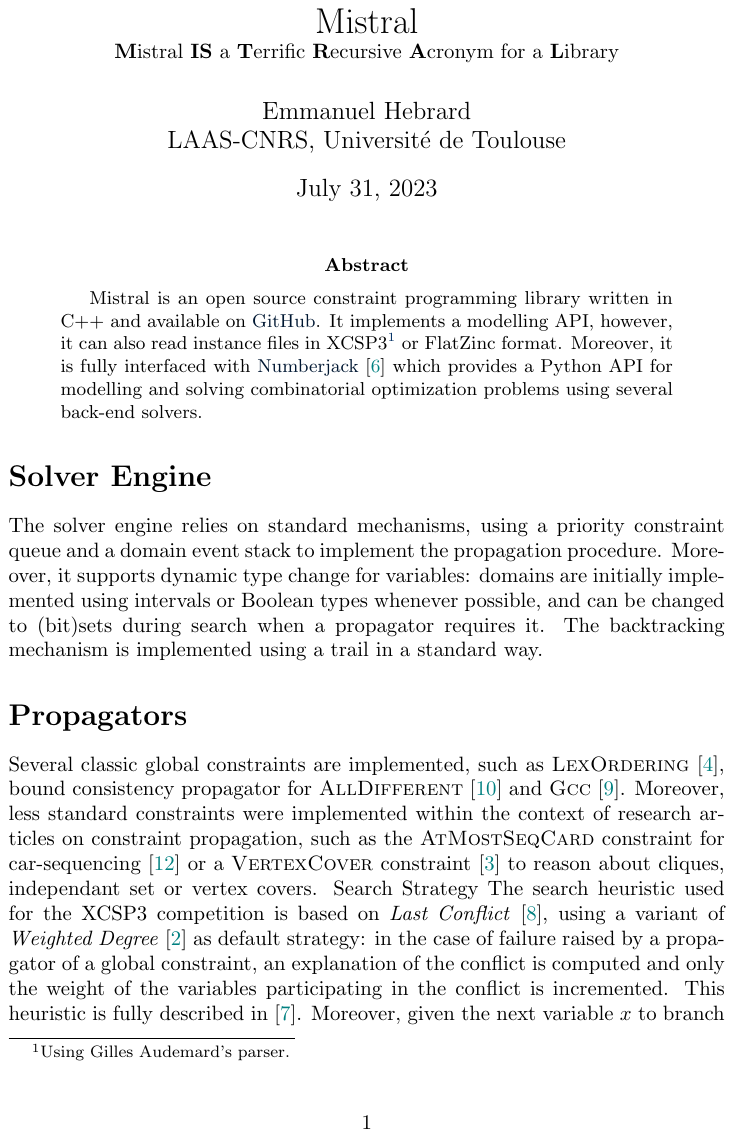}
\addcontentsline{toc}{section}{\numberline{}NACRE}
\includepdf[pages=-,pagecommand={\thispagestyle{plain}}]{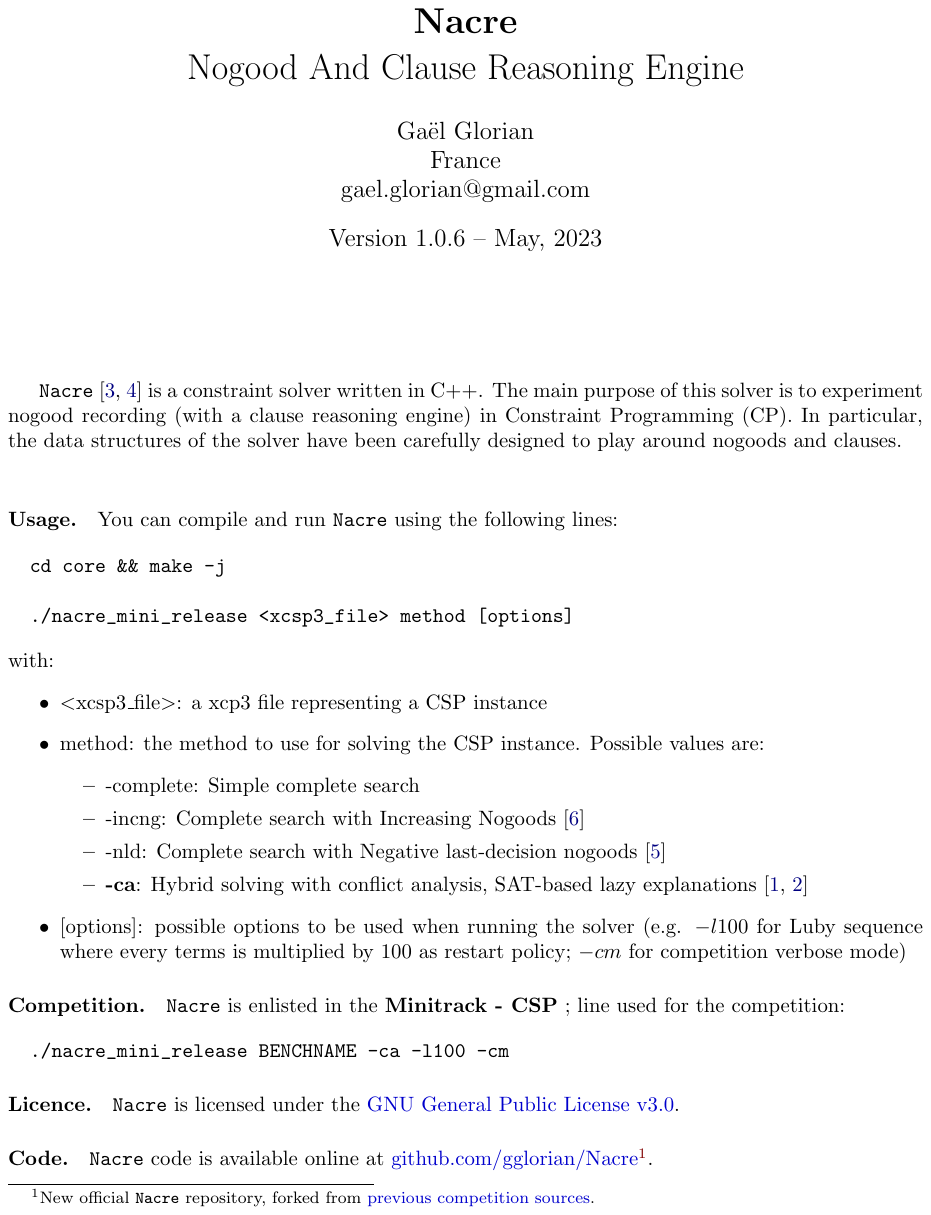}
\addcontentsline{toc}{section}{\numberline{}Picat}
\includepdf[pages=-,pagecommand={\thispagestyle{plain}}]{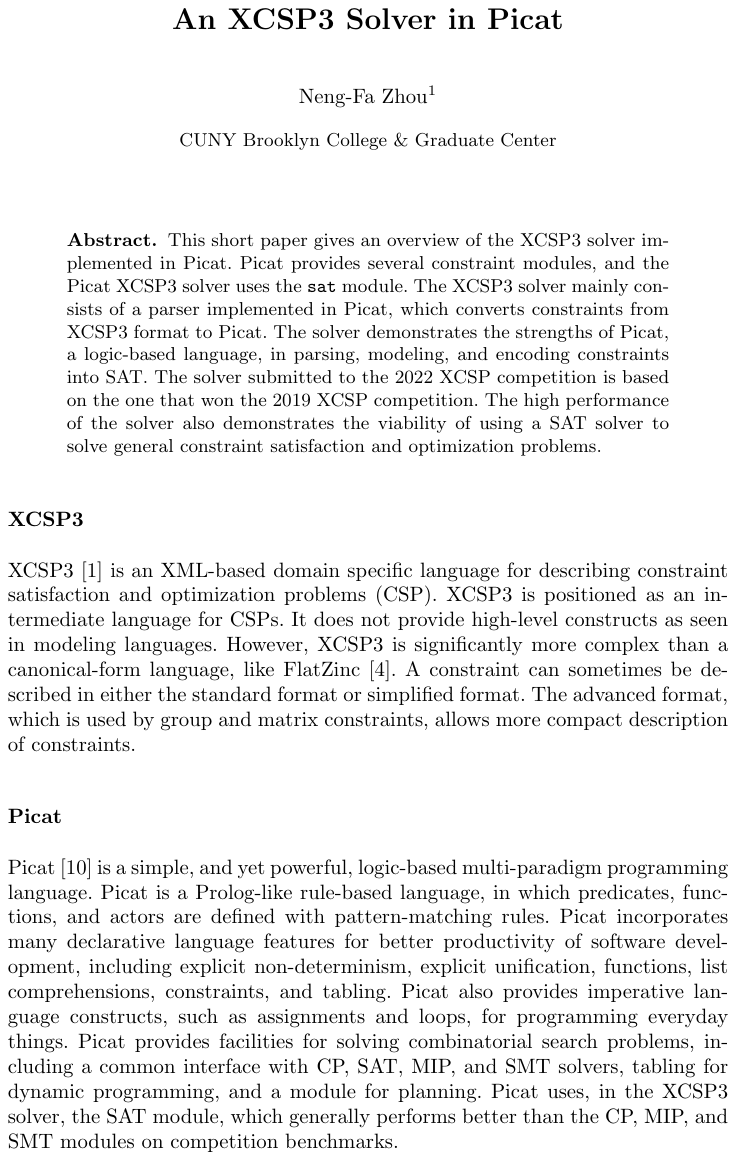}
\addcontentsline{toc}{section}{\numberline{}RBO}
\includepdf[pages=-,pagecommand={\thispagestyle{plain}}]{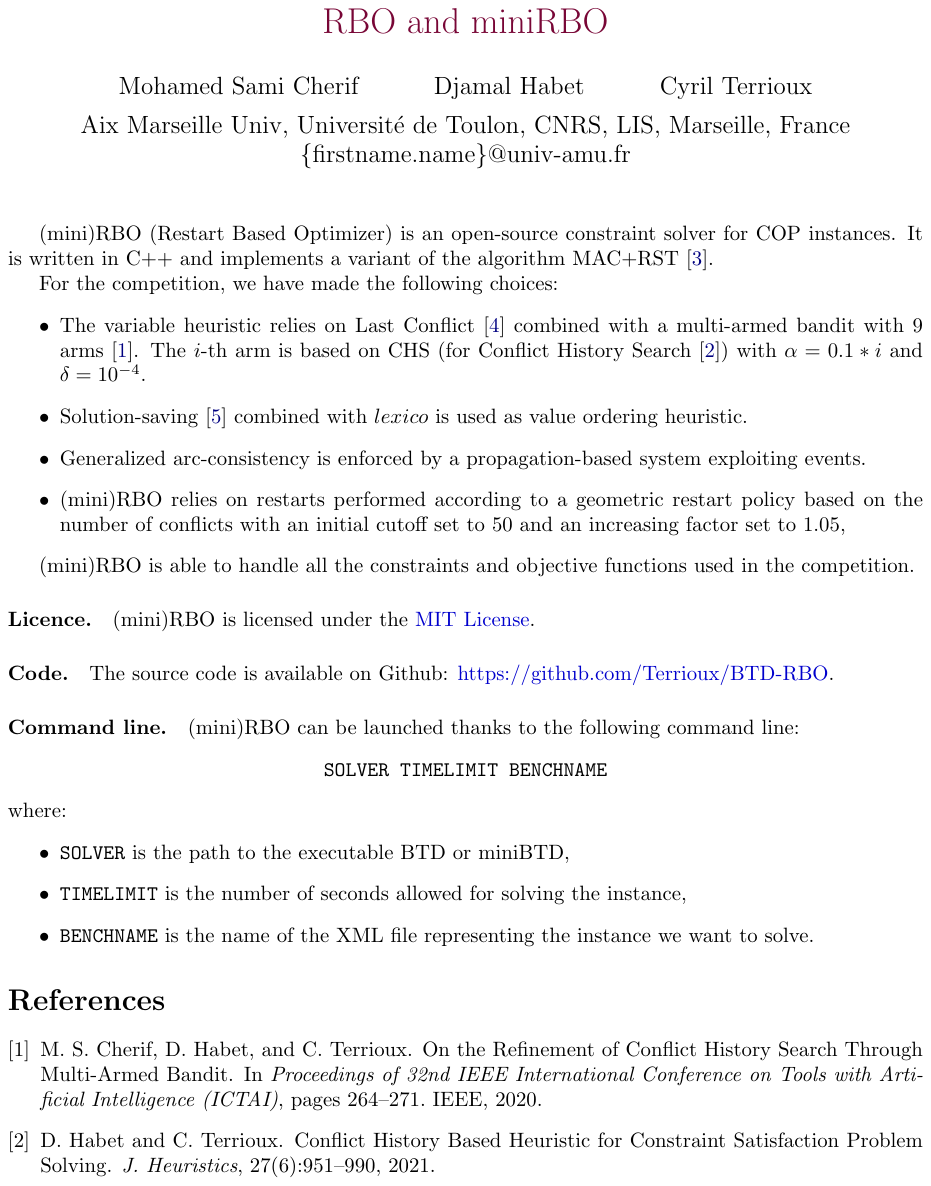}
\addcontentsline{toc}{section}{\numberline{}Sat4j-CSP-PBj}
\includepdf[pages=-,pagecommand={\thispagestyle{plain}}]{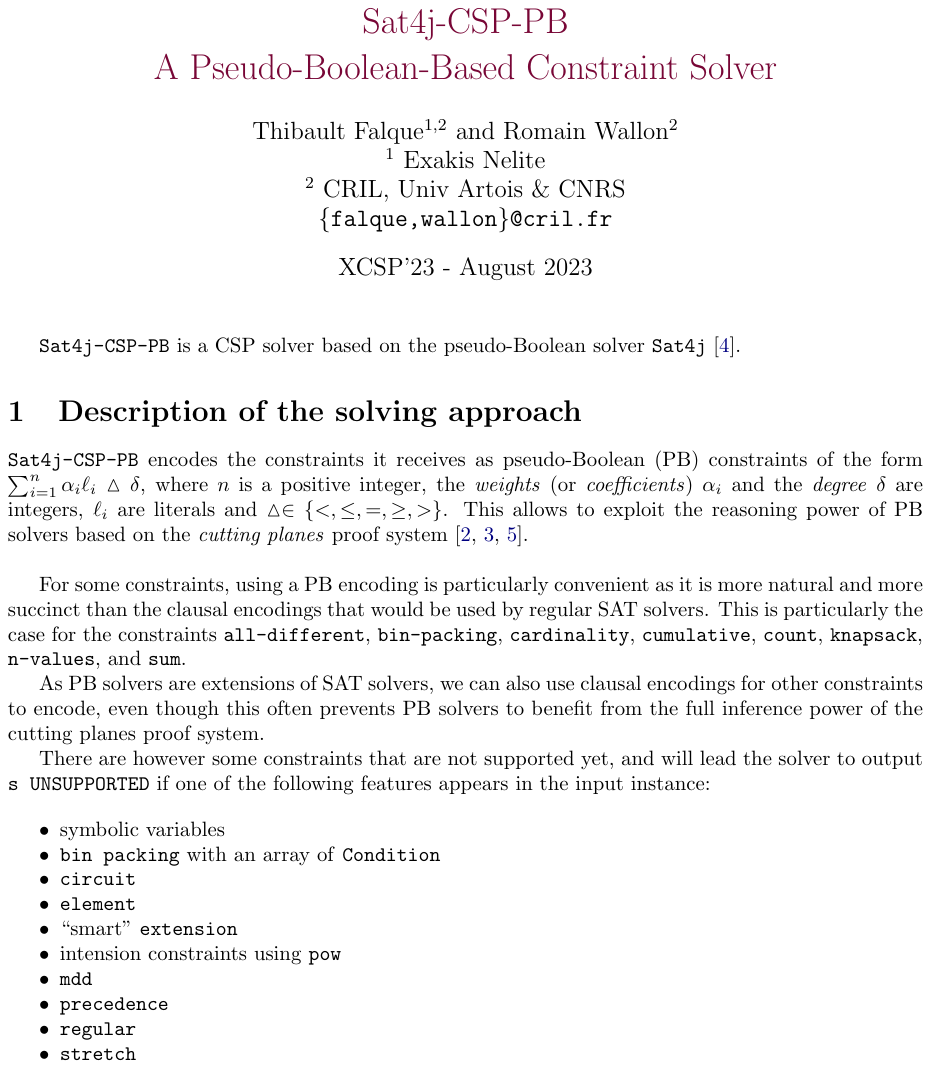}
\addcontentsline{toc}{section}{\numberline{}SeaPearl}
\includepdf[pages=-,pagecommand={\thispagestyle{plain}}]{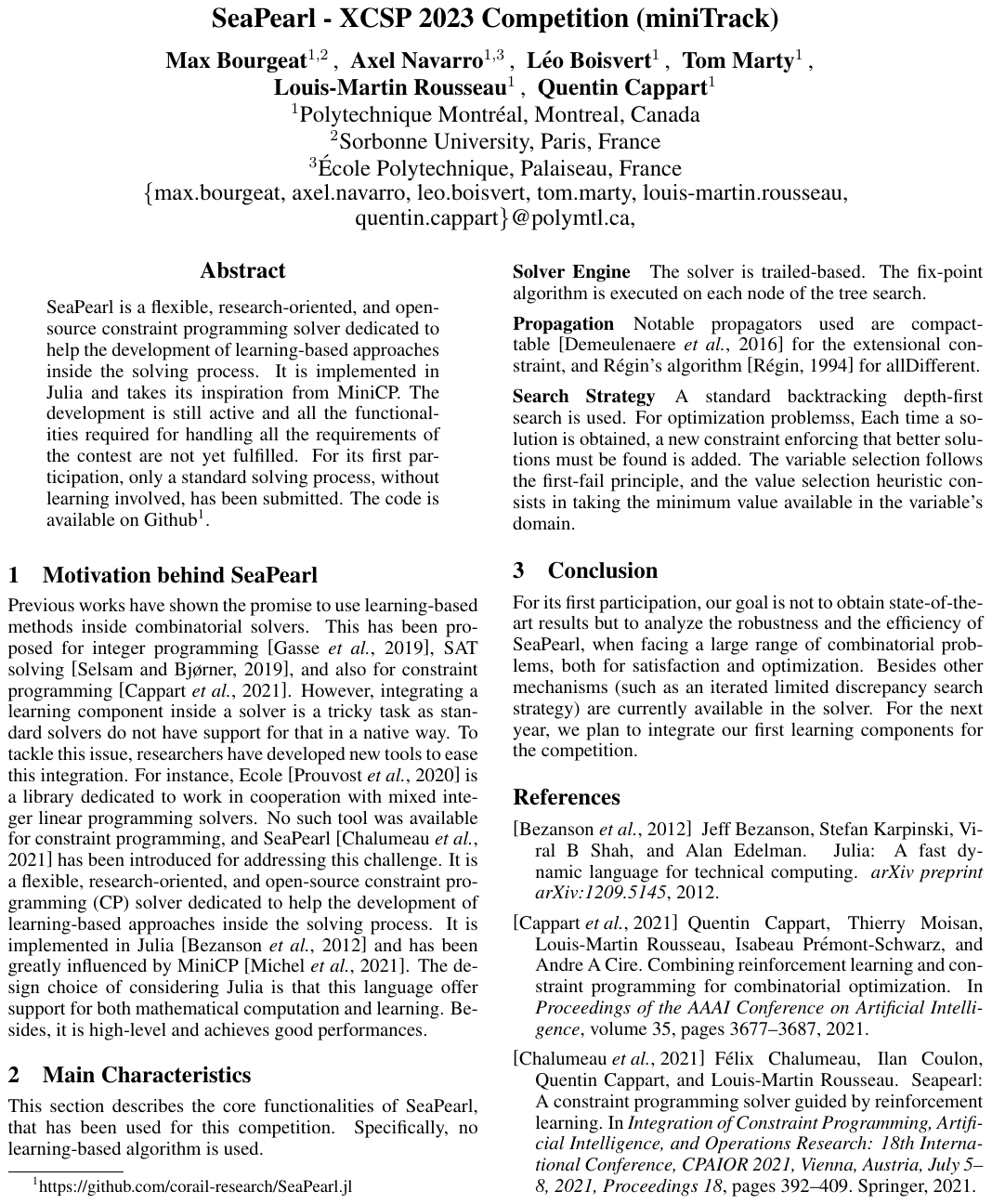}
\addcontentsline{toc}{section}{\numberline{}toulbar2}
\includepdf[pages=-,pagecommand={\thispagestyle{plain}}]{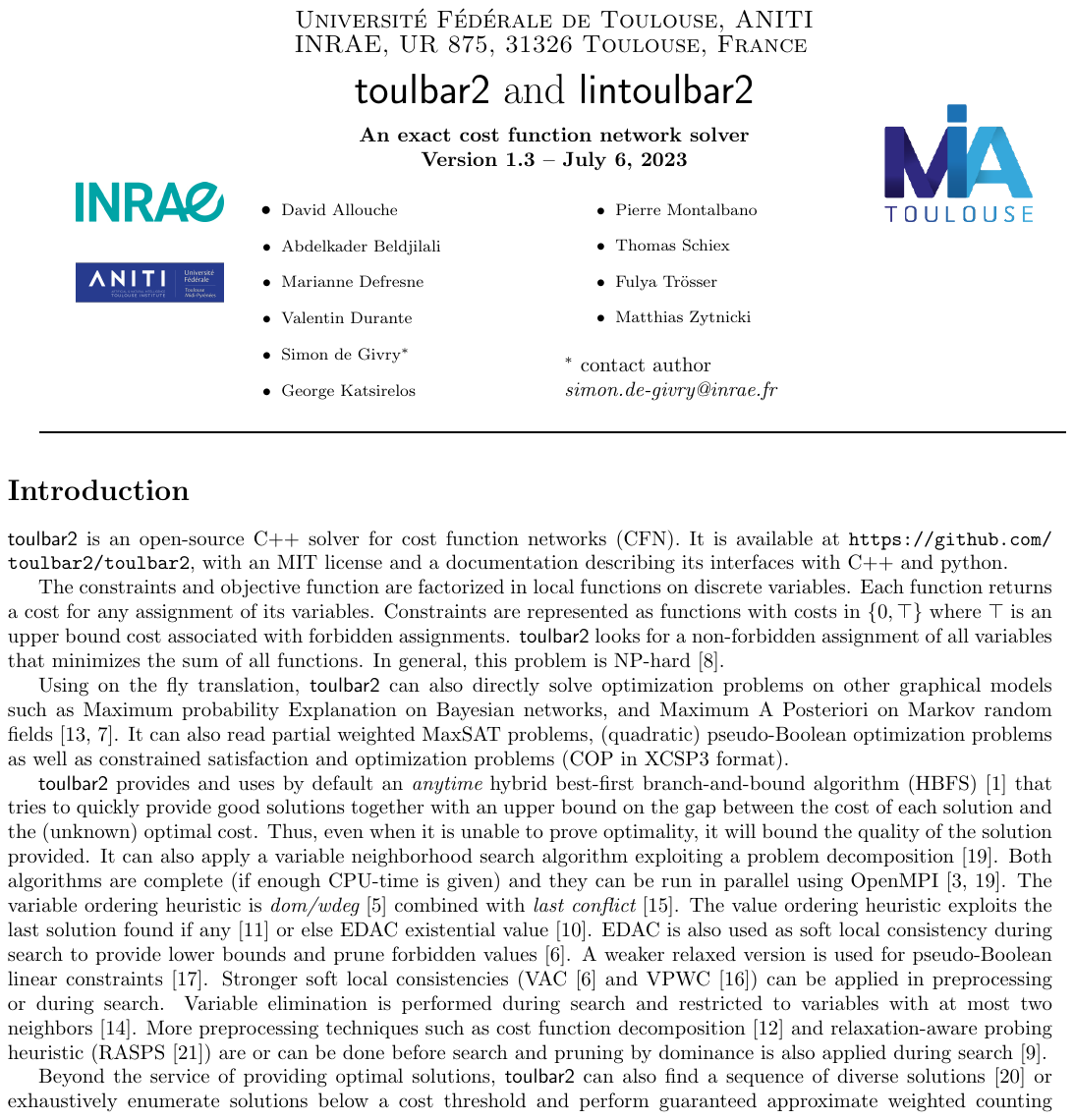}

%\includepdf[nup=1x1, frame, scale=0.9,pages=-,pagecommand={\thispagestyle{plain}}]{BTD_12.pdf}

\chapter{Results}

In this chapter, rankings for the various tracks of the \x3 Competition 2023 are given.
%We also make a few general comments about the results.
Importantly, remember that you can find all detailed results, including all traces of solvers at \href{https://www.cril.univ-artois.fr/XCSP23/}{https://www.cril.univ-artois.fr/XCSP23/}.

\section{Context}

\bigskip
Remember that the tracks of the competition are given by Table \ref{tab:anysolver} and Table \ref{tab:minisolver}.

\begin{table}[h!]
\begin{center}
\begin{tabular}{cccc} 
\toprule
\textcolor{dred}{\bf Problem} &  \textcolor{dred}{\bf Goal} &  \textcolor{dred}{\bf Exploration} &  \textcolor{dred}{\bf Timeout} \\
\midrule
CSP  & one solution & sequential & 40 minutes \\
COP  & best solution & sequential & 40 minutes \\
Fast COP  & best solution & sequential & 4 minutes \\
// COP  & best solution & parallel & 40 minutes \\
\bottomrule
\end{tabular}
\end{center}
\caption{Standard Tracks. \label{tab:anysolver}}
\end{table}

\begin{table}[h!]
\begin{center}
\begin{tabular}{cccc} 
\toprule
\textcolor{dred}{\bf Problem} &  \textcolor{dred}{\bf Goal} &  \textcolor{dred}{\bf Exploration} &  \textcolor{dred}{\bf Timeout} \\
\midrule
Mini CSP  & one solution & sequential & 40 minutes \\
Mini COP  & best solution & sequential & 40 minutes \\
%COP  & optimal solution & sequential & 40 minutes \\
\bottomrule
\end{tabular}
\end{center}
\caption{Mini-Solver Tracks. \label{tab:minisolver}}
\end{table}

\noindent Also, note that:

\begin{itemize}
\item the cluster was provided by CRIL and is composed of nodes with two quad-cores (Intel Xeon CPU E5-2637 v4 @ 3.50GHz, each equipped with 64 GiB RAM).
\item Hyperthreading was disabled.
\item Each solver was allocated a CPU and 64 GiB of RAM, independently from the tracks.
\item Timeouts were set accordingly to the tracks through the tool \texttt{runsolver}:
  \begin{itemize}
    \item sequential solvers in the fast COP track were allocated 4 min of CPU time and 12~min of Wall Clock time,
    \item other sequential solvers were allocated 40 min of CPU time and 120 min of Wall Clock time,
    \item parallel solvers were allocated 160 min of CPU time and 120 min %(VÉRIFIER LES TIMEOUTS POUR CETTE TRACK et ajouter une note sur le timeout tel que compté dans la compétition)
      of Wall Clock time.
  \end{itemize}
\item The selection of instances for the Standard tracks was composed of 200 CSP instances and 250~COP instances.
\item The selection of instances for the Mini-solver tracks was composed of 150 CSP instances and 155~COP instances.
\end{itemize}
%\noindent Concerning the selection of instances, we end up with: 
%\begin{itemize}
%\item Standard tracks: 236 CSP and 346 COP instances
%\item Mini-solver tracks: 176 CSP and 188 COP instances 
%\end{itemize}

\paragraph{About Scoring.}
The number of points won by a solver $S$ is decided as follows:
\begin{itemize}
\item for CSP, this is the number of times $S$ is able to solve an instance, i.e., to decide the satisfiability of an instance (either exhibiting a solution, or indicating that the instance is unsatisfiable)
\item for COP, this is, roughly speaking, the number of times $S$ gives the best known result, compared to its competitors.
More specifically, for each instance $I$:
\begin{itemize}
\item if $I$ is unsatisfiable, 1 point is won by $S$ if $S$ indicates that the instance $I$ is unsatisfiable, 0 otherwise,
\item if $S$ provides a solution whose bound is less good than another one (found by another competiting solver), 0 point is won by $S$,
\item if $S$ provides an optimal solution, while indicating that it is indeed the optimality, 1 point is won by $S$,
\item if $S$ provides (a solution with) the best found bound among all competitors, this being possibly shared by some other solver(s), while indicating no information about optimality: 1 point is won by $S$ if no other solver proved that this bound was optimal, $0.5$ otherwise.
%\item if $S$ provides (a solution with) the best found bound, possibly shared by some other solver(s), , with no indication hile no solver with no solver being able to out indicating that it is indeed optimality, 1 point is won by $S$ is no other solver proved optimality, $0.5$ otherwise.
\end{itemize}
\end{itemize}

%Notice that for COP, another viewpoint is the number of times a solver can give the best known answer (optimality or best known bound); it is then given between parentheses in some tables.

\paragraph{Off-competition Solvers.} Some solvers were run while not officially entering the competition: we call them {\em off-competition} solvers.
\ace is one of them because its author (C. Lecoutre) conducted the selection of instances, which is a very strong bias.
Also, when two variants (by the same competiting team) of a same solver compete in a same track, only the best one is ranked (and the second one considered as being off-competition).
This is why, for example, Fun-sCOP-glue was considered as off-competition in the CSP track.

\section{Rankings}

Recall that, concerning ranking, two new rules are used when necessary:
\begin{itemize}
\item In case a team submits the same solver to both the main track and the mini-track for the same problem (CSP or COP), 
the solver will be ranked in the mini-track only if the solver is not one of the three best solvers in the main track.
\item In case several teams submit variations of the same solver to the same track, only
the team who developed the solver and the best other team with that solver will
be ranked (possibly, a second best other team, if the jury thinks that it is relevant)
\end{itemize}

\bigskip
\noindent The algorithm used in practice for establishing the ranking is:
\begin{enumerate}
\item first, off-competition solvers are discarded (this is the case for ACE in 2023)
\item second, in mini-tracks, solvers that are ranked 1st, 2nd or 3rd in the corresponding main track are discarded
 (for example, in 2023, this is the case for Choco and Mistral in the mini COP track)
\item  third, any worse variation of the same solver (submitted by the same team) is discarded 
  (for example, this is the case of lintoulbar2 and sa4j-resolution in the 2023 miniCOP track)
\end{enumerate}

\bigskip
Here are the rankings\footnote{The images of medals come from \href{https://freesvg.org/gold-medal-juhele-final}{freesvg.org}} for the 6 tracks.
\bigskip
% https://freesvg.org/bronze-medal-juhele-final
% https://freesvg.org/gold-medal-juhele-final
% https://freesvg.org/silver-medal-juhele-final

\begin{minipage}{0.4\textwidth}
\begin{center}
  \begin{tabular}{|lcp{0.0cm}p{2.1cm}|}
    \hline
     \vspace{-0.2cm} & & &  \\
    \multirow{5}{2.1cm}{{\bf {\large ~ CSP}}} & \includegraphics[scale=0.15]{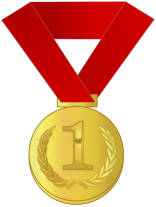} & & \vspace{-0.6cm} {\large Picat} \\
    & & & \\
    & \includegraphics[scale=0.15]{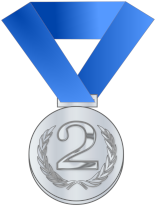} & & \vspace{-0.6cm} {\large Fun-sCOP} \\
    & & & \\
    & \includegraphics[scale=0.15]{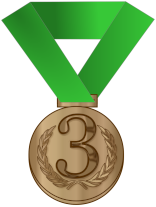} & & \vspace{-0.6cm} {\large Choco} \\
    \hline 
  \end{tabular}
\end{center}
\end{minipage} \hspace{1.4cm}
\begin{minipage}{0.4\textwidth}
\begin{center}
  \begin{tabular}{|lcp{0.0cm}p{2.1cm}|}
    \hline
    \vspace{-0.2cm} & & &  \\
    \multirow{5}{2.1cm}{{\bf {\large ~ COP}}} & \includegraphics[scale=0.15]{gold.png} & & \vspace{-0.6cm}  {\large Mistral} \\
    & & & \\
    & \includegraphics[scale=0.15]{silver.png} & & \vspace{-0.6cm} {\large Choco} \\
    & & & \\
    & \includegraphics[scale=0.15]{bronze.png} & & \vspace{-0.6cm} {\large CoSoCo} \\
    \hline
  \end{tabular}
\end{center}
\end{minipage}

\bigskip

\begin{minipage}{0.4\textwidth}  
\begin{center}
  \begin{tabular}{|lcp{0.0cm}p{2.1cm}|}
    \hline
    \vspace{-0.2cm} & & &  \\
    \multirow{5}{2.1cm}{{\bf {\large Fast COP}}} & \includegraphics[scale=0.15]{gold.png} & & \vspace{-0.6cm} {\large Choco} \\
    & & & \\
    & \includegraphics[scale=0.15]{silver.png} & & \vspace{-0.6cm} {\large Mistral} \\
    & & & \\
    & \includegraphics[scale=0.15]{bronze.png} & & \vspace{-0.6cm} {\large CoSoCo} \\
    \hline
  \end{tabular}
\end{center}
\end{minipage} \hspace{1.4cm}
\begin{minipage}{0.4\textwidth}
\begin{center}
  \begin{tabular}{|lcp{0.0cm}p{2.1cm}|}
    \hline
    \vspace{-0.2cm} & & &  \\
    \multirow{5}{2.1cm}{{\bf {\large // COP}}} & \includegraphics[scale=0.15]{gold.png} & & \vspace{-0.6cm} {\large Choco} \\
    & & & \\
    & \includegraphics[scale=0.15]{silver.png} & & \vspace{-0.6cm} {\large Toulbar2} \\
    & & & \\
    & \includegraphics[scale=0.15]{bronze.png} & & \vspace{-0.6cm} {\large ~ --}  \\
    \hline
  \end{tabular}
\end{center}
\end{minipage}

\bigskip

\begin{minipage}{0.4\textwidth}  
\begin{center}
  \begin{tabular}{|lcp{0.0cm}p{2.1cm}|}
    \hline
    \vspace{-0.2cm} & & &  \\
    \multirow{5}{2.1cm}{{\bf {\large Mini CSP}}} &\includegraphics[scale=0.15]{gold.png} & & \vspace{-0.6cm} {\large Exchequer} \\
    & & & \\
    & \includegraphics[scale=0.15]{silver.png} & & \vspace{-0.6cm} {\large miniBTD} \\
    & & & \\
    & \includegraphics[scale=0.15]{bronze.png} & & \vspace{-0.6cm} {\large Nacre} \\
    \hline
  \end{tabular}
\end{center}
\end{minipage} \hspace{1.4cm}
\begin{minipage}{0.4\textwidth}
\begin{center}
  \begin{tabular}{|lcp{0.0cm}p{2.1cm}|}
    \hline
    \vspace{-0.2cm} & & &  \\
    \multirow{5}{2.15cm}{{\bf {\large Mini COP}}} & \includegraphics[scale=0.15]{gold.png} & & \vspace{-0.6cm} {\large Toulbar2} \\
    & & & \\
    & \includegraphics[scale=0.15]{silver.png} & & \vspace{-0.6cm} {\large Exchequer} \\
    & & & \\
    & \includegraphics[scale=0.15]{bronze.png} & & \vspace{-0.6cm} {\large Sat4j-both} \\
    \hline
  \end{tabular}
\end{center}
\end{minipage}

\bibliographystyle{plain} %alpha}
%\bibliography{./globalBiblio}

\begin{thebibliography}{10}

\bibitem{AFGJMN_conjure}
O.~Akgün, A.~Frisch, I.~Gent, C.~Jefferson, I.~Miguel, and P.~Nightingale.
\newblock Conjure: Automatic generation of constraint models from problem
  specifications.
\newblock {\em Artificial Intelligence}, 310, 2022.

\bibitem{AE_efficient}
A.~El Amraoui and M.~Elhafsi.
\newblock An efficient new heuristic for the hoist scheduling problem.
\newblock {\em Computers \& Operations Research}, 67:184--192, 2016.

\bibitem{B_note}
J.~Beasley.
\newblock A note on solving large p-median problems.
\newblock {\em European Journal of Operational Research}, 21:270--273, 1985.

\bibitem{BHHKW_slide}
C.~Bessiere, E.~Hebrard, B.~Hnich, Z.~Kiziltan, and T.~Walsh.
\newblock {SLIDE}: A useful special case of the {CARDPATH} constraint.
\newblock In {\em Proceedings of ECAI'08}, pages 475--479, 2008.

\bibitem{D_binary}
M.~De Biazi.
\newblock Binary puzzle is {NP-complete}.
\newblock Technical report, ResearchGate, 2012.
\newblock
  \url{https://www.researchgate.net/publication/243972408_Binary_Puzzle_is_NP-complete}.

\bibitem{BLAPxcsp3}
F.~Boussemart, C.~Lecoutre, G.~Audemard, and C.~Piette.
\newblock {\em {XCSP3}: An Integrated Format for Benchmarking Combinatorial
  Constrained Problems}.
\newblock Technical Report. v3.1 on CoRR,
  \href{https://arxiv.org/pdf/1611.03398.pdf}{arXiv:1611.03398}, 2016--2022.
\newblock 241 pages.

\bibitem{BLAP_xcsp3core}
F.~Boussemart, C.~Lecoutre, G.~Audemard, and C.~Piette.
\newblock {\em {XCSP3}-core: A Format for Representing Constraint
  Satisfaction/Optimization Problems}.
\newblock Technical Report. v3.1 on CoRR,
  \href{https://arxiv.org/pdf/2009.00514.pdf}{arXiv:2009.00514}, 2020--2022.
\newblock 105 pages.

\bibitem{CHPTV_how}
Y.~Carissan, D.~Hagebaum-Reignier, N.~Prcovic, C.~Terrioux, and A.~Varet.
\newblock How constraint programming can help chemists to generate benzenoid
  structures and assess the local aromaticity of benzenoids.
\newblock {\em Constraints}, 27(3):192--248, 2022.

\bibitem{CGS_solving}
V.~Coppé, X.~Gillard, and P.~Schaus.
\newblock Solving the constrained single-row facility layout problem with
  decision diagrams.
\newblock In {\em Proceedings of CP'22}, pages 14:1--14:18, 2022.

\bibitem{DPS_optimizing}
J.~Dickerson, A.~Procaccia, and T.~Sandholm.
\newblock Optimizing kidney exchange with transplant chains: theory and
  reality.
\newblock In {\em Proceedings of AAMAS'12}, pages 711--718, 2012.

\bibitem{DAD_soccer}
R.~Duque, A.~Arbelaez, and J.-F. Diaz.
\newblock {CP} and {MIP} approaches for soccer analysis.
\newblock {\em Journal of industrial and management optimization},
  15(4):1535--1564, 2019.

\bibitem{DDa_sabio}
R.~Duque, J.-F. Díaz, and A.~Arbelaez.
\newblock {SABIO}: An implementation of mip and cp for interactive soccer
  queries.
\newblock In {\em Proceedings of CP'16}, pages 575--583, 2016.

\bibitem{GS_hybrid}
L.~Di Gaspero and A.~Schaerf.
\newblock Hybrid local search techniques for the generalized balanced academic
  curriculum problem.
\newblock In {\em Proceedings of HM'08}, pages 146--157, 2008.

\bibitem{GHS_simple}
S.~Gay, R.~Hartert, and P.~Schaus.
\newblock Simple and scalable time-table filtering for the cumulative
  constraint.
\newblock In {\em Proceedings of CP'15}, pages 149--157, 2015.

\bibitem{GS_lns}
X.~Gillard and P.~Schaus.
\newblock Large neighborhood search with decision diagrams.
\newblock In {\em Proceedings of IJCAI'22}, pages 4754--4760, 2022.

\bibitem{HSW_item}
V.~Houndji, P.~Schaus, and L.~Wolsey.
\newblock The item dependent stockingcost constraint.
\newblock {\em Constraints}, 24(2):183--209, 2019.

\bibitem{JMMT_peg}
C.~Jefferson, A.~Miguel, I.~Miguel, and A.~Tarim.
\newblock Modelling and solving english peg solitaire.
\newblock In {\em Proceedings of CPAIOR'03}, 2003.

\bibitem{KSSZ_mixed}
S.~Kreter, A.~Schutt, P.~Stuckey, and J.~Zimmermann.
\newblock Mixed-integer linear programming and constraint programming
  formulations for solving resource availability cost problems.
\newblock {\em European Journal of Operational Research}, 266(2):472--486,
  2018.

\bibitem{ace}
C.~Lecoutre.
\newblock {\em {ACE}, a generic constraint solver}.
\newblock Technical Report. v2.1 on CoRR,
  \href{https://doi.org/10.48550/arXiv.2302.05405}{arXiv.2302.05405}, 2023.

\bibitem{LS_pycsp3}
C.~Lecoutre and N.~Szczepanski.
\newblock {\em {PyCSP3}: Modeling Combinatorial Constrained Problems in
  Python}.
\newblock Technical Report. v2.1 on CoRR,
  \href{https://arxiv.org/pdf/2009.00326.pdf}{arXiv:2009.00326}, 2020--2022.
\newblock 155 pages.

\bibitem{M_kep}
V.~Mak-Hau.
\newblock On the kidney exchange problem: cardinality constrained cycle and
  chain problems on directed graphs: a survey of integer programming
  approaches.
\newblock {\em Journal of Combinatorial Optimization}, 33(1):35--59, 2017.

\bibitem{MP_optimal}
M.D. Moffitt and M.E. Pollack.
\newblock Optimal rectangle packing: A meta-{CSP} approach.
\newblock In {\em Proceedings of ICAPS'06}, pages 93--102, 2006.

\bibitem{RCP23}
S.~Roussel, T.~Polacsek, and A.~Chan.
\newblock Assembly line preliminary design optimization for an aircraft.
\newblock In {\em Proceedings of CP'23}, 2023.

\bibitem{SSV_optimal}
A.~Schutt, P.~Stuckey, and A.~Verden.
\newblock Optimal carpet cutting.
\newblock In {\em Proceedings of CP'11}, pages 69--84, 2011.

\bibitem{S_dominoes}
H.~Simonis.
\newblock Dominoes as a constraint problem.
\newblock Technical report, CiteSeerX, 2007.

\bibitem{SO_search}
H.~Simonis and B.~O'Sullivan.
\newblock Search strategies for rectangle packing.
\newblock In {\em Proceedings of CP'08}, pages 52--66, 2008.

\bibitem{SBHW_progressive}
B.~Smith, S.~Brailsford, P.~Hubbard, and P.~Williams.
\newblock The progressive party problem: Integer linear programming and
  constraint programming compared.
\newblock {\em Constraints}, 1(1-2):119--138, 1996.

\bibitem{S_teaching}
P.~Szeredi.
\newblock Teaching constraints through logic puzzles.
\newblock In {\em Proceedings of CSCLP'03}, pages 196--222, 2003.

\bibitem{HM_constraint}
P.~van Hentenryck and L.~Michel.
\newblock {\em Constraint-based local search}.
\newblock MIT Press, 2005.

\bibitem{WY_new}
M.~Wallace and N.~Yorke-Smith.
\newblock A new constraint programming model and solving for the cyclic hoist
  scheduling problem.
\newblock {\em Constraints}, 25(3-4):319--337, 2020.

\end{thebibliography}

\end{document}